%% file: main.tex
\pdfoutput=1

\documentclass[11pt]{article}

\usepackage[final]{acl}

\usepackage{times}
\usepackage{latexsym}

\usepackage[T1]{fontenc}

\usepackage[utf8]{inputenc}

\usepackage{microtype}

\usepackage{inconsolata}

\usepackage{amsmath}
\usepackage{graphicx}
\usepackage{tabularx}
\usepackage{cleveref}
\usepackage{tablefootnote}
\usepackage{algorithm}
\usepackage{algpseudocode}
\usepackage{mdframed}
\usepackage{xcolor}
\def\statements/{\texttt{statements}}
\def\Statements/{\texttt{Statements}}
\def\statement/{\texttt{statement}}
\def\Statement/{\texttt{Statement}}

\title{Statements: Universal Information Extraction from Tables
with Large Language Models for ESG KPIs}

\author{
 \textbf{Lokesh Mishra\textsuperscript{1}},
 \textbf{Sohayl Dhibi\textsuperscript{1}},
 \textbf{Yusik Kim\textsuperscript{2}},
 \\
 \textbf{Cesar Berrospi Ramis\textsuperscript{1}},
\textbf{Shubham Gupta\textsuperscript{2}},
 \textbf{Michele Dolfi\textsuperscript{1}},
 \textbf{Peter Staar\textsuperscript{1}},
\\
\\
 \textsuperscript{1}IBM Research Zurich, Säumerstrasse 4, Rüschlikon, Switzerland,
 \\
 \textsuperscript{2}IBM Research Paris-Saclay,  2 Rue d'Arsonval, Orsay, France
\\
\small \texttt{[mis, ceb, dol, taa]@zurich.ibm.com} 
\\
\small \texttt{[sohayl.dhibi, yusik.kim, shubham.gupta1]@ibm.com}
}

\begin{document}
\maketitle
\begin{abstract}
Environment, Social, and Governance (ESG) KPIs assess an organization's performance on issues such as climate change, greenhouse gas emissions, water consumption, waste management, human rights, diversity, and policies. ESG reports convey this valuable quantitative information through tables.Unfortunately, extracting this information is difficult due to high variability in the table structure as well as content. We propose \Statements/, a novel domain agnostic data-structure for extracting quantitative facts and related information. We propose translating tables to \statements/ as a new supervised deep-learning universal information extraction task. We introduce SemTabNet -- a dataset of over 100K annotated tables. Investigating a family of T5-based Statement Extraction Models, our best model generates \statements/ which are 82\% similar to the ground-truth (compared to baseline of $21\%$). We demonstrate the advantages of \statements/ by applying our model to over 2700 tables from ESG reports. The homogeneous nature of \statements/ permits exploratory data analysis on expansive information found in large collections of ESG reports.
\end{abstract}

\section{Introduction}
\label{sec:introduction}
It is invaluable to assess mankind's impact on climate. Climate change related information is often published in so-called  ``Environment, Social, and Governance (ESG)'' reports. Corporations report valuable quantitative data regarding their efforts to improve their impact on environment, working conditions, and company culture in these ESG reports \citep{bingler_how_2022, schimanski_bridging_2024}. 

Like most technical documents, ESG reports present their key information in tables, making table understanding and information extraction (IE) an important problem \citep{mishra_esg_2024}. This problem becomes further complicated due to the large variety and diversity of tabular representations used in these reports. Despite efforts to standardize these reports, this diversity makes the task of extracting information from these documents extremely challenging (see Appendix Fig. \ref{fig:complex_table} for an example table).

Large Language Models (LLMs) have turned out to be excellent tools for IE, due to their ability to parse, understand, and reason over textual data \citep{openai_gpt-4_2023, touvron_llama_2023}. This, in combination with their in-context learning ability, makes them excellent for IE from text \citep{brown_language_2020}. This approach breaks down when applying the same techniques on tables \citep{zhu_converting_2021}.

\begin{figure}[]
    \centering
    \includegraphics[width=\linewidth]{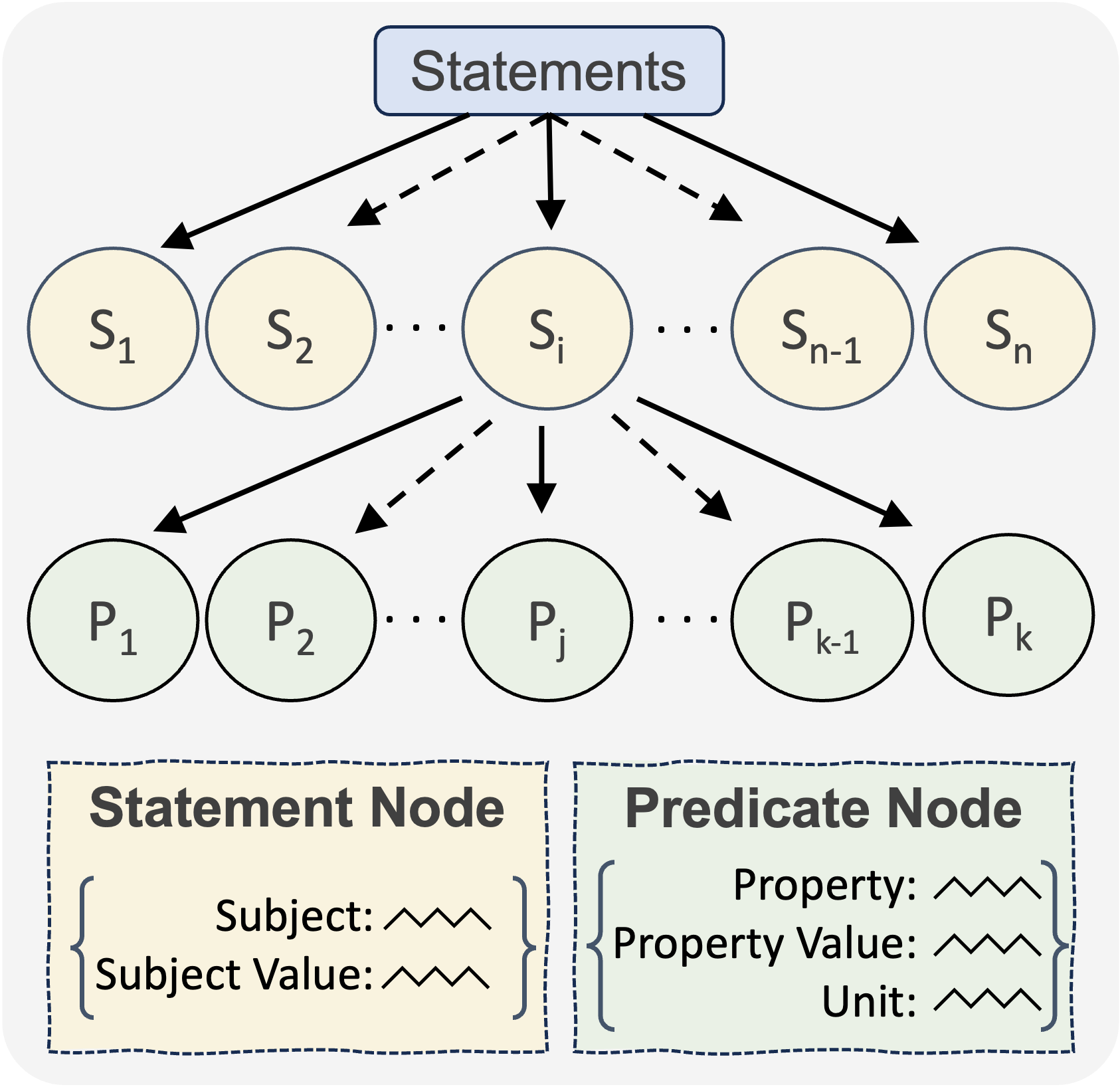}
    \caption{The knowledge model of \Statements/ represented as a tree. From the root node, individual statements emerge as branches. Associated with each individual statement node are the leaf predicate nodes.}
    \label{fig:knowledge_model}
\end{figure}

\begin{figure*}
    \centering
    \includegraphics[width=\linewidth]{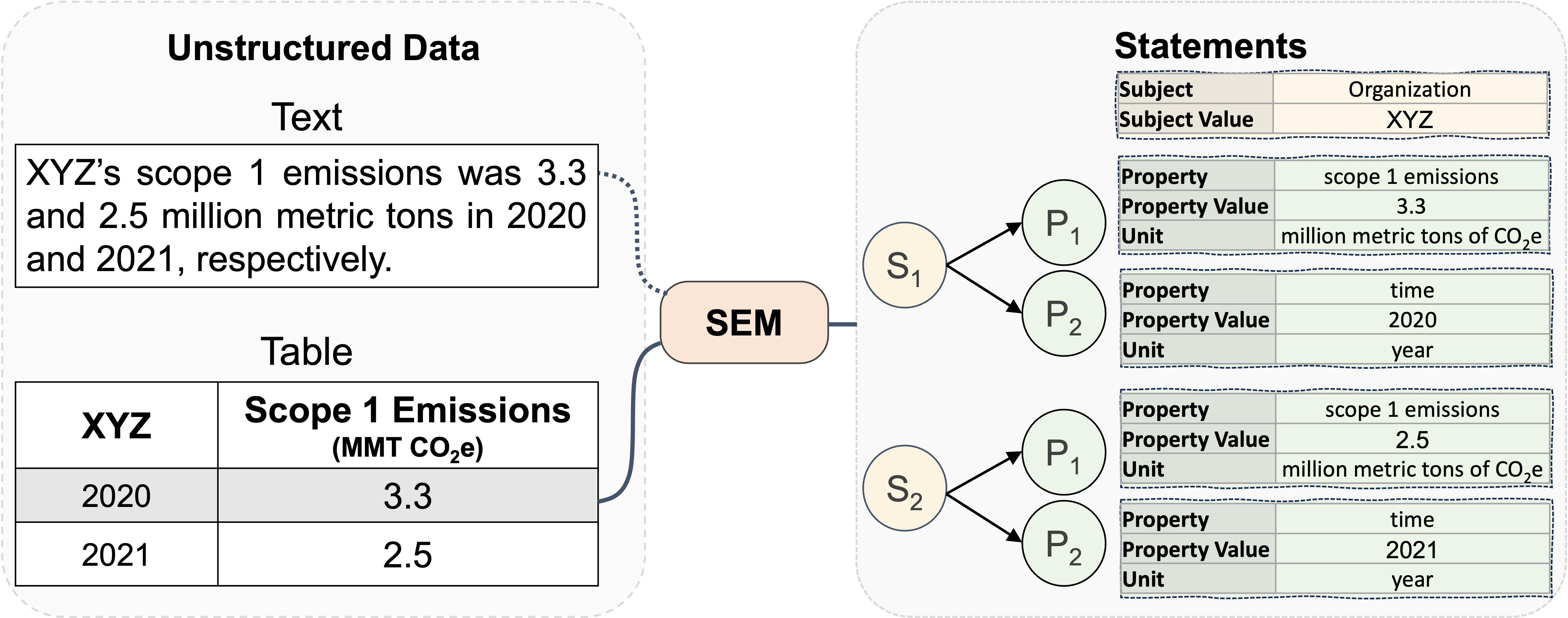}
    \caption{A diagram explaining the framework introduced in this paper. We fine-tune LLMs on the task of `Statement Extraction' leading to a family of ``\textbf{S}tatement \textbf{E}xtraction \textbf{M}odels'' (SEM). Quantitative facts are extracted from heterogenous unstructured data (only tables in this paper) and stored as Statements.}
    \label{fig:text_table_to_statements}
\end{figure*}

In this paper, we present a general approach for universal IE from tables. Universal IE involves named entity recognition and relationship extraction among other tasks. To this end, we propose a new tree-like data structure, called `\Statement/', which can combine multiple (named) entities and (n-ary) relations (Fig. \ref{fig:knowledge_model}). It allows us to represent information in a homogeneous domain agnostic fashion. 
A \statement/ tree can contain content from different subjects, allowing for universal IE approach to tables across multiple domains.
With the introduction of \statements/, the IE problem from tables becomes a \textit{translation problem} which we call \emph{`statement extraction'} -- translating the original table into a set of statements. ESG reports, to this day, are manually analyzed by consultancy firms and professional organisations \citep{henisz_five_2019}. With our proposed statement extraction, this process can now be fully automated.

To evaluate our model generated statements, we propose a novel application of the well-established Tree Edit Distance \cite{pawlik_tree_2016}. We propose Tree Similarity Score ($t_s$) for measuring the similarity between two trees. 
As baseline,  we experiment with in-context learning using state-of-the-art LLMs like Mistral \cite{jiang_mistral_2023}, Mixtral \cite{Jiang2024MixtralOE}, Llama2 \cite{touvron_llama_2023}, and Falcon \cite{almazrouei2023falcon}. These models show an average $t_s$ varying from $0\%$ to $21\%$. On the other hand, our best-performing fine-tuned T5 based model shows a $t_s$ of $82\%$.Our main contributions are:
\begin{itemize}
\setlength\itemsep{-0.5 em}
\item We introduce a new knowledge model called \Statement/ for mapping complex, irregular, and heterogeneous information to a uniform domain agnositc structure.
\item We present a new supervised deep learning universal IE task called \emph{`statement extraction'}. The fine-tuned models show significant improvement over baseline experiments providing competitive benchmarks for the community.   
\item We contribute to the field of table understanding, by providing ``SemTabNet'' a dataset containing over 100K annotated ESG tables. All cells in these tables are annotated to reflect their semantic relationship with other cells.
\item We propose Tree Similarity Score, which in a single number quantifies the quality of entities and relationships extraction in the statement.
\end{itemize}

We begin, in Sect. \ref{sec:related_work} discussing related works. In Sect. \ref{sec:statements} we explain the concept of `\Statements/' and  
present the SemTabNet dataset in Sect. \ref{sec:data}. In sect. \ref{sec:experimentalresults}, we discuss the various experiments we performed and their results. We end the paper with an application of our model on ESG reports.

\section{Related works}
\label{sec:related_work}

\citet{fang_large_2024} group the applications of deep learning methods to tables or tabular data into four broad categories. (1) Tree based methods such as gradient-boosted decision trees \citep{borisov_deep_2022} for predictions on tabular data. (2) Attention-based methods which includes developing models that learn tabular representations such as TAPAS \cite{herzig-etal-2020-tapas}, TABERT \cite{yin-etal-2020-tabert}, 
and/or fine-tuning models for downstream tasks on tabular data like fact-checking  \cite[TABFACt]{2019TabFact}, question-answering \cite{liu_tapex_2021, mishra_esg_2024}, semantic parsing \cite{yu_grappa_2020}.
(3) Regularization methods which attempts to modify model sensitivity to tabular features \cite{kadra_well-tuned_2021}. (4) Data transformation methods which aim at converting heterogeneous tabular inputs to homogeneous data, like an image \citep{sun_supertml_2019} or feature engineering \citep{Liu2020DNN2LRIF}. 

Another class of problem which is similar to the data transformation approach is (generative) information extraction (IE) which involves adopting LLMs to generate structural information from an information source.
Recent studies have found that LLMs can also perform universal IE \cite{kardas_axcell_2020, paolini_structured_2020, wang_deepstruct_2022, wang_instructuie_2023}. 

In a universal IE task, a model is trained to generate desirable structured information $y$, given a pre-defined schema $s$, and information source $x$ \cite{lu_unified_2022}. Using pre-trained language models, \citet{wang_ielm_2022} perform IE in two steps: argument extraction and predicate extraction. Based on this, they introduced a text-based open IE benchmark. \citet{wang_zero-shot_2021} presented DeepEx for extracting structured triplets from text based data. \citet{wang_deepstruct_2022} demonstrate that pre-training models on task-agnostic corpus lead to performance improvement on tasks like IE, entity recognition, etc. However, these approaches are limited to textual data.

\citet{bai_schema-driven_2024} have shown that LLMs can perform IE on tabular data when prompted with a table and a relevant extraction schema. Their approach is based on a human-in-the-loop in-context learning.  A domain-expert is necessary for producing robust extraction schema, which instructs the model to generate structured records from a table. This strongly limits the adaptability of their approach to different domains. Although limited to text, \cite{lu_unified_2022} also propose a schema-driven universal IE system. They use a structure extraction language which generates structural schema prompt which guides the model in its IE tasks. 

As we show, the \statements/ data structure removes several limitations of previous universal IE approaches and is applicable to `wild' heterogenous information sources.

\section{Definition of \Statements/}
\label{sec:statements}

The \statements/ data structure aims to homogenize data coming from complex, irregular, heterogeneous information source (text or tables). At its core, the \statements/ data structure is a tree structure (\cref{fig:knowledge_model}). From the root of the tree, we have `subject'-nodes, which contain information regarding the `subject' and the `subject-value' keys. From each subject-node, there are one or more predicate nodes, which define the `property', `property-value', and `unit' keys. Each predicate node carries an atomic piece of quantitative information.

The \statement/ knowledge model can be applied to both text and tables. In Fig. \ref{fig:text_table_to_statements}, we show the same \statements/ structure which could be obtained from a text or a corresponding table. As such, the \statements/ structure is not bound only to tables, however, it shows its usefulness particularly when normalising information from heterogeneous tables. The details of how we create trees are presented with examples in \cref{appendix:ted}.

The tree structure of \statements/ allows us to quantify, with a single number, the transformation of information from a table. This is accomplished by computing the Tree Similarity Score (based on the Tree Editing Distance (TED)~\citet{pawlik_tree_2016, schwarz_new_2017}) between predicted and ground-truth \statements/. TED is defined as the minimum-cost sequence of node operations that transform one tree into another. 
Like the Levenshtein distances on strings \citep{1966SPhD...10..707L}, TED involves three kinds of operations: node insertions, deletions, and renaming. The cost of each operation can be freely defined, which makes this metric both flexible and powerful. Two trees are exactly same when their tree similarity score is 100\%. To ensure high quality statement extraction, we setup robust TED costs such that minor differences can lead to poor tree similarity scores. In \cref{appendix:treesimilarityscore}, we demonstrate tree similarity score with some examples.

It is also instructive to look at the edit types which converted the predicted statements into ground-truth statements. For this, we measure the ratio of edit type to the total number of edits. 
We find that the ratio of insertions and ratio of deletions carries the information about the structural similarity of two trees. If the model predicted too few nodes, the ratio of insertions will be high. Correspondingly, if the statements from the model's prediction has too many nodes, the deletion ratio dominates. If two trees are structurally similar, then the ratio of both insertion and deletions is low. In this case, the edits are dominated by renaming.

While tree-based metrics are sensitive to both entity and relationship extraction, we also would like to understand the ability of a Statement Extraction Model (SEM) to extract entities alone \footnote{Here, `entity' refers to the values of attributes in a statement. For example, `scope 1 emissions' is an entity from the statement shown in \cref{fig:text_table_to_statements}.}. For this, we concatenate all the predicate nodes in a statement. We create sets of values corresponding to: subject, subject value, property, property value, unit. We count true positives when an entity is found in both the sets from model prediction and ground truth. True negatives are counted when an entity is present only in the ground truth set and false positives when the entity is present only in the predicted set. Based on these, we measure the standard accuracy, recall and F1 measures.

\section{SemTabNet: \Statements/ Data}
\label{sec:data}

\begin{figure*}
    \centering
    \includegraphics[width=\linewidth]{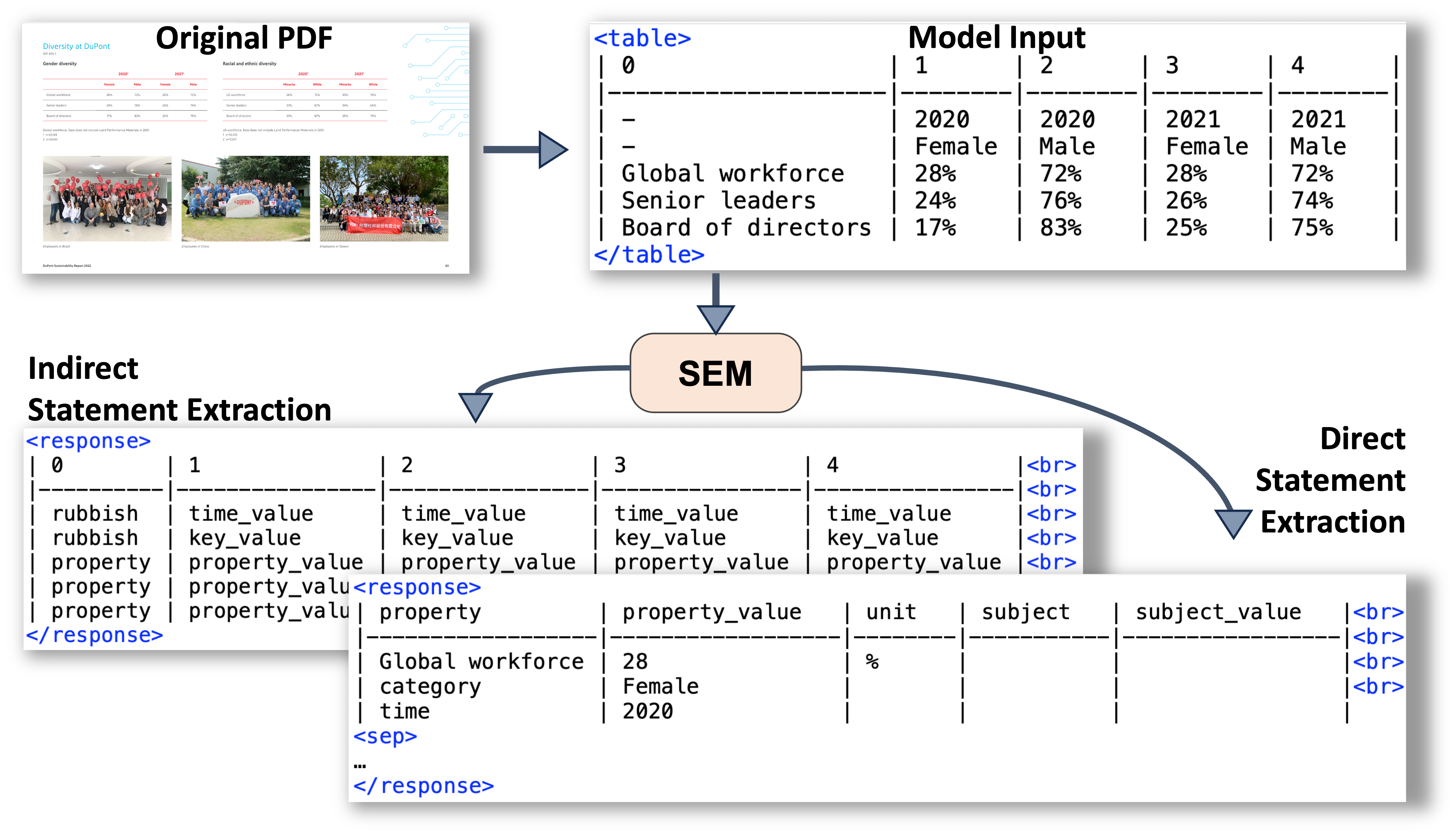}
    \caption{Input and output for the task of ``Statement Extraction''. \textit{Top Left:} Page from an ESG report containing tables. \textit{Top Right:} One of the table, from the same page, prepared as markdown for model input. \textit{Bottom Left:} Model output for the task of indirect statement extraction. \textit{Bottom Right:} Model output for the task of direct statement extraction.}
    \label{fig:model_input_output}
\end{figure*}

There are many large data sets of annotated tables which suffer from two major limitations: (1) they focus on understanding table structure only i.e. demarcating table headers from table content, and (2) contain little diversity in shape, size, and complexity of the table. Tables found in ESG reports are of high complexity with little common underlying pattern. In this work, we advance deep learning on table understanding by annotating the content of the table and annotating complex tables. 

We used the Deep Search toolkit~\footnote{Available via: \href{https://ds4sd.github.io}{https://ds4sd.github.io}.} to collect over 10K ESG reports from over 2000 corporations. Deep Search crawled these PDF reports, converted them into machine readable format, and provided this data along with the metadata of each report in json format. 

We compiled a list of important keywords which capture many important concepts in ESG reports (see \cref{appendix:esgkeywords}). Next, we select only those tables which have some relevance with the keywords. For this we used the following conditions: the ROUGE-L precision (longest common sub-sequence) score between raw data and keywords must be greater than 0.75 and there must be quantitative information in the table.

We need a strategy for understanding the content of a table and extracting statements from it. After manually observing hundreds of table, we decided a two step approach to prepare our ground-truth data. First, we classify all the cells in a table based on the semantic meaning of their content into 16 categories which helps us in constructing statements. For each table, this step creates a `labels-table' with the same shape and structure as the original, but the cells of this labels-table only contain category labels (see \cref{fig:model_input_output}). Secondly, we create a program which reads both the labels-table and the original table and extracts statements in a rule-based approach. The algorithm is described in \cref{alg:statement_extraction}.
The 16 labels are:  
\begin{itemize}
    \setlength\itemsep{-0.5em}
    \item Property, Property Value
    \item Sub-property
    \item Subject, Subject Value 
    \item Unit, Unit Value
    \item Time, Time Value
    \item Key, Key Value
    \item Header 1, Header 2, Header 3
    \item Empty, Rubbish
\end{itemize}

During annotation, all cells are mapped to one of the above labels. For cells which contain information pertaining to more than one label, we pick the label which is higher in our ordered list of labels. So ``Revenue (US\$)'', is labelled as \texttt{property}. 
The `property' and `sub-property' cells always have associated `property value' cell(s). The `header' cells never have an associated value and often divide the table into smaller sections. Empty cells are labelled `empty'. When a table contain unnecessary parts due to faulty table recovery or non-quantitative information. We label such cells as `rubbish'. When a property/property value pair carries supplementary information, those cells are annotated as `key'/`key values'.

Additionally, we observed that most tables can be reasonably classified into three baskets: simple, complex, and qualitative. There are simple tables whose structure cannot be further subdivided into any smaller table. There are complex tables whose structure can be further divided into multiple smaller tables. Finally, there are qualitative tables (like table of contents) which contain little valuable information for our endeavour. 

We collected about 2,800 tables and found $\sim 20\%$ had simple layout, $\sim 20\%$ had complex layout (composed of multiple simpler tables arranged hierarchically), and $\sim 60\%$ were qualitative. We discarded all qualitative tables from any further analysis. To ensure that our data is not biased towards either simple or complex tables, we manually annotated all the cells of 569 simple tables and 538 complex tables. In total, we annotated 1,107 tables (84,890 individual cells) giving rise to 42,982 statements. 

Due to the nature of our strategy, one can extract statements from tables either directly in a zero shot manner (direct SE) or by predicting cell labels and then using the rule-based approach to construct statements (indirect SE) (see Fig. \ref{fig:model_input_output}. We have experimented with both approaches. 

We further augmented the annotated tables to create a large training data. We shuffle the rows and columns of tables corresponding to property-values to create new augmented tables, while keeping their contents the same. While this is straightforward for simple tables, special care was taken for complex tables such that only rows/columns which belonged together within a category were shuffled. The maximum number of augmented tables emerging from the shuffling operations was limited to 130, leading to over 120K tables. To promote further research and development, we open source this large dataset of semantic cell annotations as SemTabNet\footnote{Links for code and data, respectively:\\ \href{https://github.com/DS4SD/SemTabNet}{https://github.com/DS4SD/SemTabNet} \href{https://huggingface.co/datasets/ds4sd/SemTabNet}{https://huggingface.co/datasets/ds4sd/SemTabNet}}. Table \ref{tab:table_counts} shows the data counts in SemTabNet. 
\input{table_counts}

\section{Experiments \& Results}
\label{sec:experimentalresults}

Fig \ref{fig:model_input_output} presents Statement Extraction as a supervised deep learning task. Due to the nature of how tables are annotated (see \cref{sec:data}), it is possible to train models for statement extraction statements both directly and indirectly. We consider the following three seq2seq experiments: (1) \textit{SE Direct}: the model is presented with an input table as markdown in a prompt. The model generates the tabular representation of the resulting statements as markdown. (2) \textit{SE Indirect 1D}: In this experiment, the model input is the individual table cell contents. For a table with $n$ cells, we predict $n$ labels sequentially (hence, 1D) and then use this information to construct statements. Individual cell labels predicted by the model are stitched together to form the labels table, which is then used to construct the predicted statement by using our rule-based algorithm. (3) \textit{SE Indirect 2D}: As opposed to SE Indirect 1D, in this experiment, we predict the cell labels of all cells in a table simultaneously. The entire table, as markdown, is input to the model (hence 2D) and the model generates the labels table, as markdown. Using the rule-based algorithm, the predicted labels table is converted into predicted statements.

We use six special tokens, which allow us to control and parse model output.
\begin{itemize}
    \setlength\itemsep{-0.5em}
    \item Input table start token: \texttt{<table>}
    \item Input table stop token: \texttt{</table>}
    \item Output start token: \texttt{<response>}
    \item Output stop token: \texttt{</response>}
    \item Newline token: \texttt{<br>}
    \item Separate list item token: \texttt{<sep>}
\end{itemize}
This allows us to parse the predicted statements from a LLM. Once successfully parsed, the output statements can be trivially converted from one representation to another. This is crucial because we compare model predicted statements with ground truth by converting statements into a tree structure. These tokens are added to the tokenizer vocabulary before fine-tuning any model.

\input{table_result}

Since the nature of these tasks naturally fits the paradigm of sequence-to-sequence models, we fine-tune T5 models \cite{raffel_exploring_2020}. T5 models are encoder-decoder transformer architecture models which are suitable for many sequence-to-sequence tasks. In our experiments, we  train T5 variants (Small, Base, Large, and 3B) to create a family of Statement Extraction Models (SEM).

In our training data for tables, the input token count is less than 512 for $50\%$ of the data, and it is less than 1024 for $90\%$ of the data. Thus, except where mentioned, we train T5 models (small, base, large) with context windows of 512 and 1024, and T5-3b with context window of 512. All models are fine-tuned in a distributed data parallel (DDP) manner simultaneously across 4 GPU devices (Nvidia A100-40GB for T5-Small, T5-Base, T5-Large and NVIDIA A100-80GB for T5-3B). Additionally, the largest possible batch size was used for all models. The batch size is impacted by factors like model size, GPU memory, and context window. In turn it affects the number of epochs we can fine-tune in a reasonable time. 

For all tasks, we stop the fine-tuning process either after 500,000 steps or after 7 days. We use the AdamW optimizer with $\beta_1 = 0.9$ and $\beta_2 = 0.999$. All models are trained with a maximum learning rate of $5 \times 10^{-4}$. There is a warm-up phase of 1000 steps in which the learning rate increases linearly from $10^{-10}$ to $5 \times 10^{-4}$. After another 1000 steps, the learning rate is exponentially decayed until it reaches its lowest value of $10^{-6}$, where it remains until the end of the training. 

Table \ref{tab:table_result} presents the key results of our experiments. For each table, we evaluate the statements predicted by the model (directly or indirectly) against the ground truth statements. For each task and each model therein, we present the averaged tree similarity score ($t_s$) (measuring entity \& relationship extraction) and the averaged F1 score (measuring entity extraction). Also present are the averaged ratios of tree edit types, which helps us understand $t_s$. For all reported values, assuming a normal distribution, the standard error of the mean is below $5 \times 10^{-5}$ and the 99\% confidence interval for all values is about $\sim 0.1\%$.

\textbf{Baseline Experiments}:
For baseline experiments, several state of the art LLMs were tested for their in-context learning ability. In the prompt, we show the model an example of direct statement extraction (1-shot), followed by a test table. 
 
The models produce statements in markdown format, which are evaluated against ground truth statements. The average tree similarity score across 1100 annotated tables varies from $0\%$ for Falcon40b to $20\%$ for Mixtral (8$\times$7b models). For entity extraction, Llama2-13b performed the best with an average F1 score of 38. Not all outputs generated by the model were in correct markdown format. 
Minor changes in the prompt were found to create vast differences in the quality of extracted statements. In \cref{appendix:prompt}, we show examples of the prompt and the model output for some cases.

\textbf{Statement Extraction Indirect 1D}:
All models trained on this task have context window of 512. Their performance tends to scale with model size. 
These models can learn to extract entities, but relationship extraction is difficult. 
For SEM-T5-small, the ratio of insertion is $\approx 98\%$ which means that the predicted statements does not have enough nodes. 

\textbf{Statement Extraction Indirect 2D}:
All models trained on this task perform well on entity extraction with average F1 scores of over $95\%$. 
The highest performing model is the SEM-T5-3b (512) with an average tree similarity score of $81.76 \%$. 

\textbf{Statement Extraction Direct}:Based on tree similarity score, most models show poor performance in direct SE. The best performing model is SEM-T5-base with a context window of 1024. It gets an average F1 score of 76.99\% and an average tree similarity score of only 11\%. To understand, why these models performs so poorly on direct SE, we look at the ratio of tree edits. 

We note that the ratio of deletions for all models in this task is close to 0. On the other hand, the ratio of insertions for all models is high (from 88\% to 98\%). This suggests that the statement trees produced by these models is missing vast number of nodes compared to the ground truth. In fact, perusing the model output shows that while the output is of high quality, it contains significantly less nodes than ground truth statements.

\begin{figure*}[ht]
    \centering
    \includegraphics[width=0.32\linewidth]{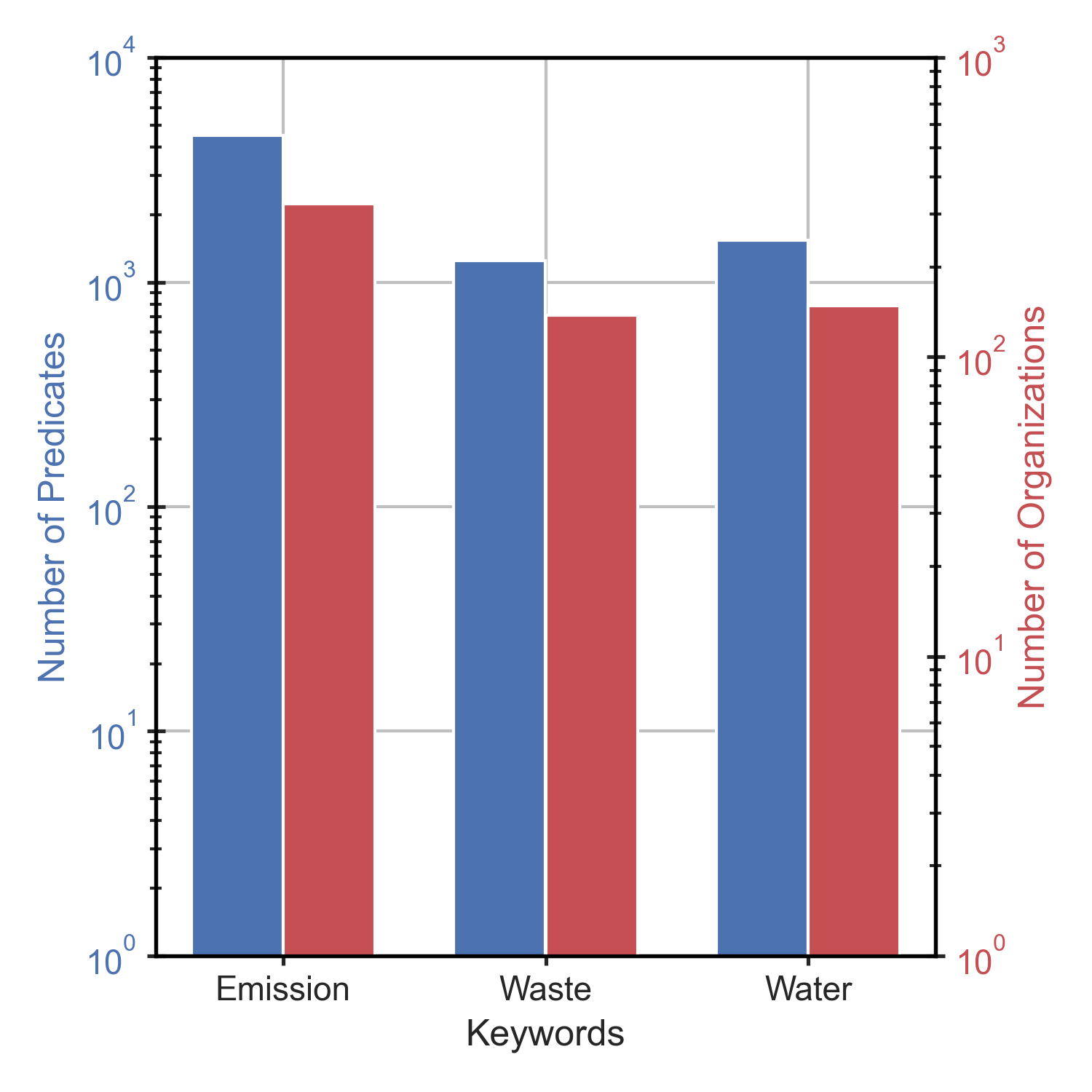}
    \includegraphics[width=0.32\linewidth]{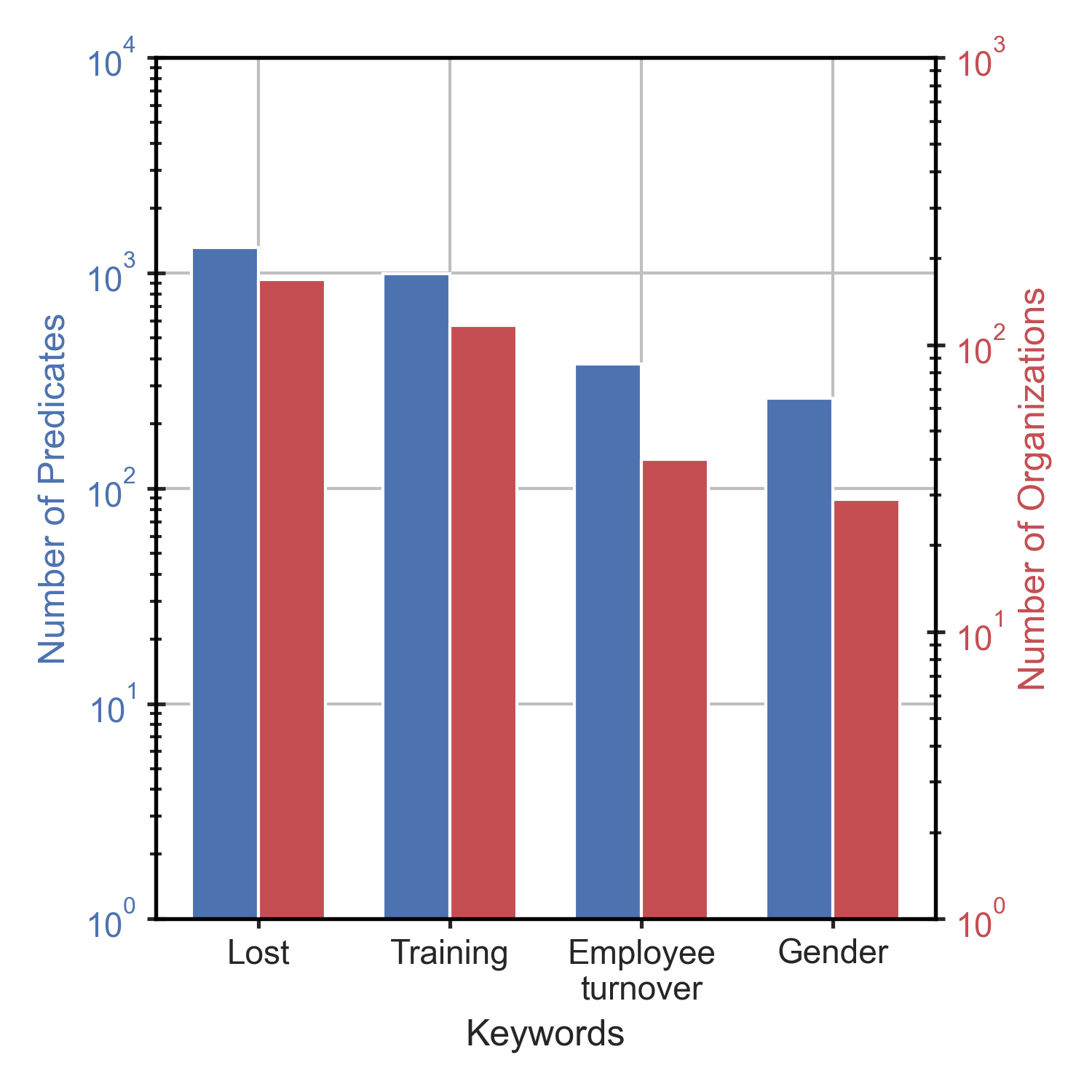}
    \includegraphics[width=0.32\linewidth]{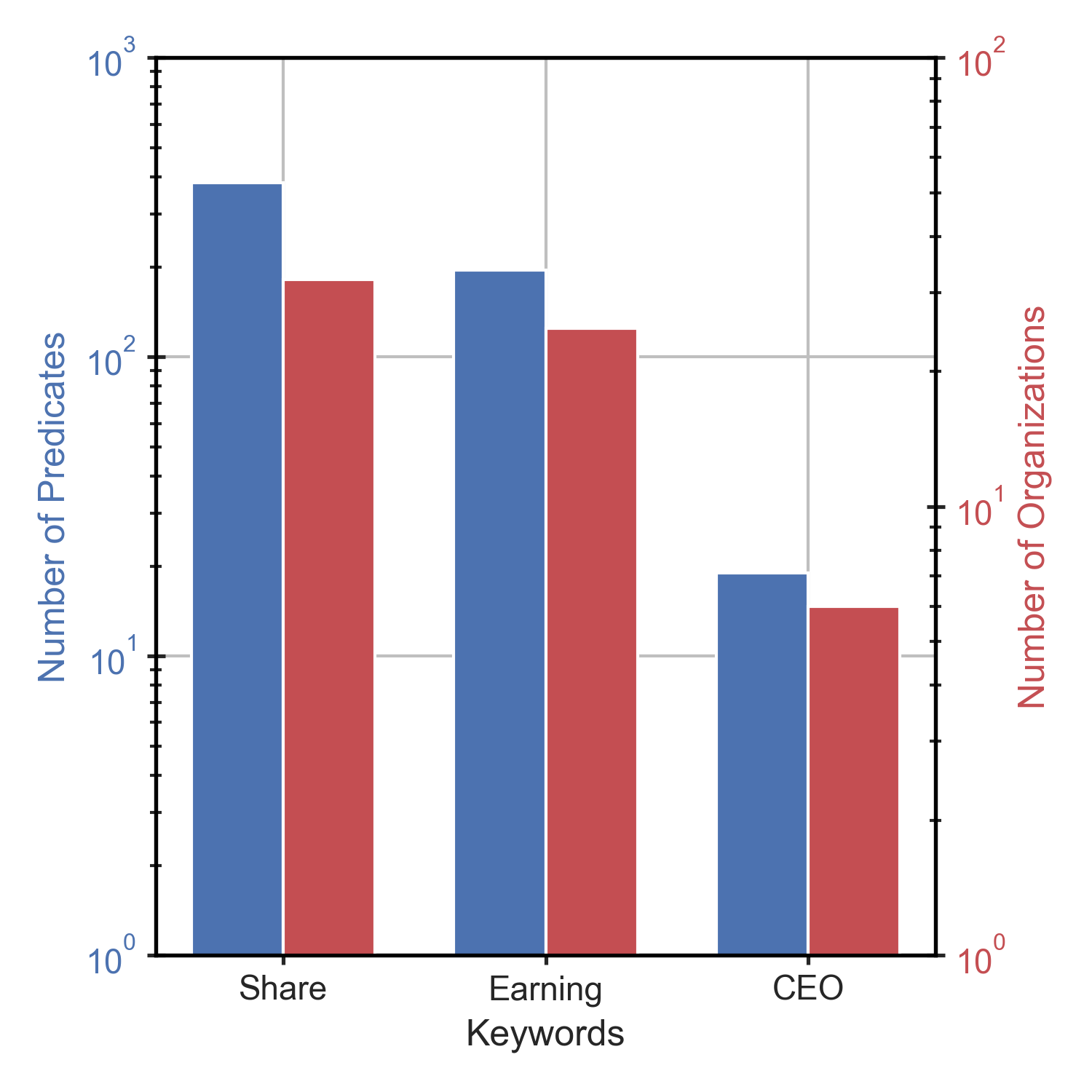}
    \\    
    \includegraphics[width=0.4\linewidth]{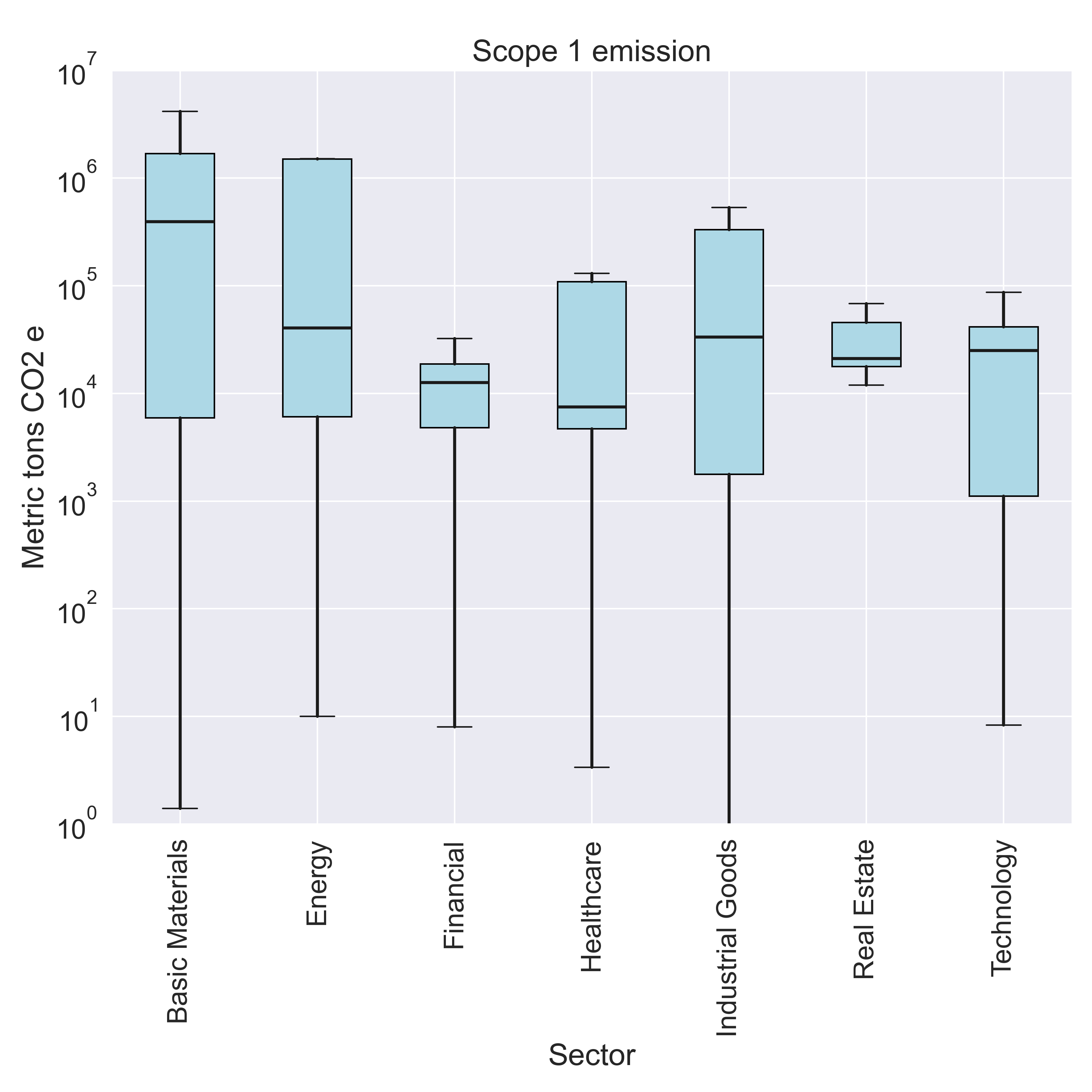} \hfill
    \includegraphics[width=0.4\linewidth]{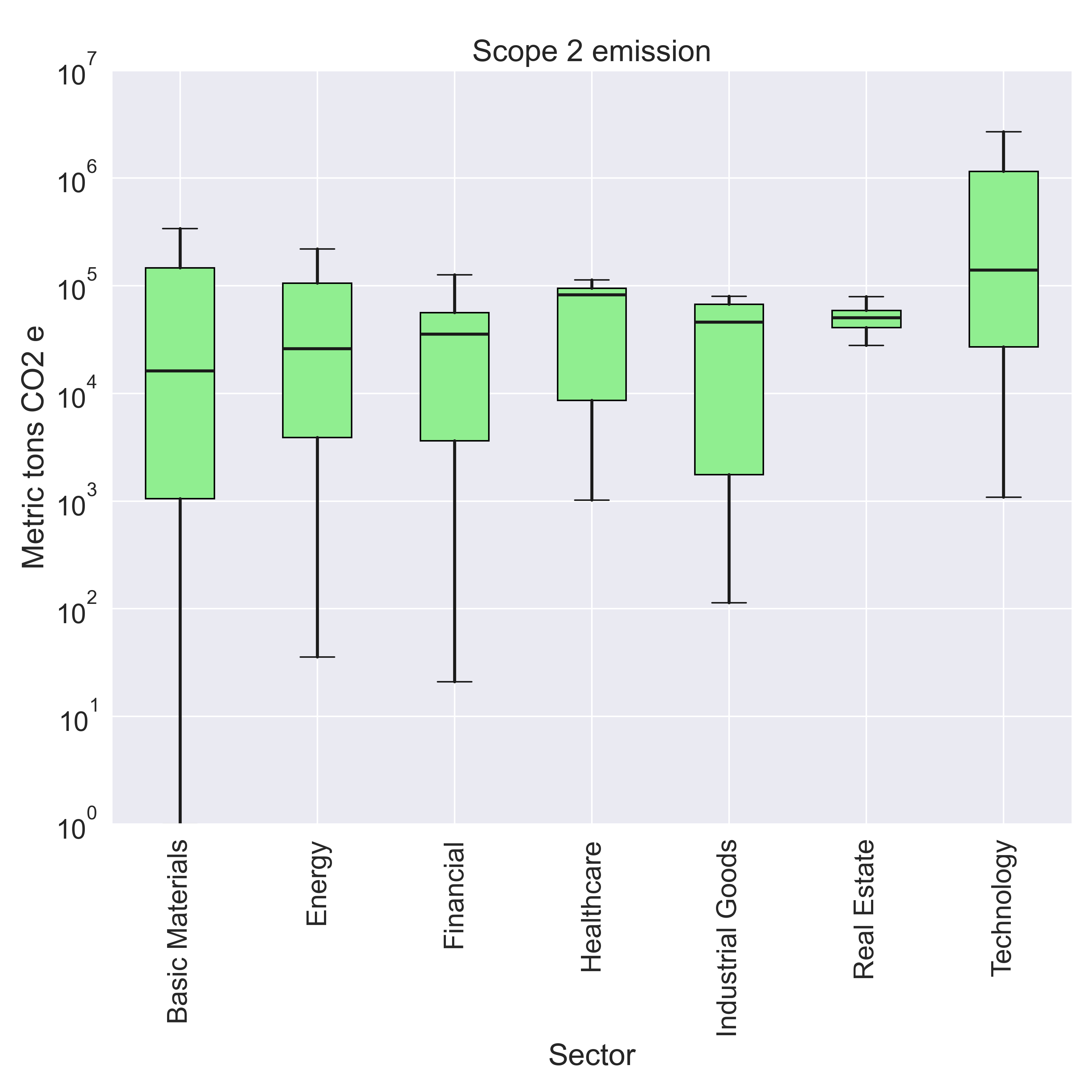}
    \caption{Exploratory data analysis of \statements/ from over 2700 Tables published in ESG reports in 2022. \textit{Top:} We searched about 50,000 predicates using keywords (shown on the x-axis) related to environment (left), social (middle), and governance (right). The plot shows the distribution of predicates and the number of organizations from this search. \textit{Bottom:} Box plot for extracted Scope 1 and Scope 2 emission values grouped by business sectors from over 300 companies across multiple years. Only sectors with more than 20 data points are included.}
    \label{fig:esg}
\end{figure*}

\textbf{Discussion}: SE Indirect 1D shows good performance on entity extraction, but performs poorly for both entity and relationship extraction. In this task, the model only sees the content of one cell at a time which makes it easy to extract entities. However, this does not allow the model to develop a strong capability to learn tabular relationships. On the other hand, SE Direct, gives poor performance on both entity extraction and relationship extraction. Direct SE expects the models to unravel a dense table into \statements/, for which they must produce many output tokens. For example, the average number of output tokens in the test data for SE direct is $5773 \pm 51$, which is significantly larger than the number of tokens for SE indirect 2D ($346 \pm 1$). Thus, direct SE is a very challenging task and might require different strategies to be executed successfully. 

SE Indirect 2D, avoids the disadvantages of both the tasks. In this case, the model sees the entire input table (has the chance to learn tabular relationships) and is only tasked with producing a labels table (can finish generation in a reasonable number of tokens). Our experiments clearly demonstrate that statement extraction via the Indirect 2D approach gives better results. This is an unexpected finding of our study, and we hope it motivates other researchers to improve zero-shot statement extraction capability.

\section{Application to ESG results}
\label{sec:application}

Due to their homogeneous structure, statements enable large-scale exploratory data analysis and data science. To demonstrate the advantage of statements over traditional tabular data science, we applied SEM-T5-large (512 SE Indirect 2D) over 2700 tables published in over 1000 ESG reports in 2022. This lead to 14,766 statements containing over 100k predicates. This dataset containing ESG related KPIs is invaluable to researchers, policy-makers, and analysts. 

We filter this large dataset to contain only those predicates with quantitative property values. This subset contains 47\,901 predicates from 601 corporate ESG reports. We search the properties in this dataset for some keywords representative of ESG KPIs. Fig. \ref{fig:esg} (top) shows the distribution of the number of predicates and the number of distinct organizations which matched our simple keyword search. For example, using `emission' as a keyword, we obtain over 4000 hits with results coming from over 300 distinct corporations. 

Fig. \ref{fig:esg} (bottoms) shows the total scope 1 emissions (left) and total scope 2 emission (right). Each box shows the distribution of emission from multiple corporations across sectors ($\sim20$ in Healthcare to $\sim100$ in Technology and Industrial Goods) containing data from several years. The data reported in the original report contained emissions in different units, which were harmonized for creating this plot. 

Since we only took a small subset of 1000 reports for this analysis, our data is incomplete and is only representative. The \statements/ dataset allows one to study how emissions from individual companies or across sectors have evolved over time. This dataset can also serve as a starting point for many other downstream applications like question-answering, fact checking, table retrieval, etc.

\section{Conclusion \& Future Works}
\label{sec:conclusions}

We have presented a novel approach to map complex, irregular, and heterogeneous information to a uniform structure, \Statements/. We presented Statement Extraction which is a new supervised deep-learning information extraction task. We contribute the field of table understanding by open-sourcing SemTabNet consisting of 100K ESG tables wherein all cells. 

Investigating three variations of the statement extraction task, we found that using a model to generate table annotations and then construct \statements/ produces best results. This approach has the advantage, that it produces homogeneous structured data with reduced hallucinations. \Statements/ are an advantageous vehicle for quantitative factual information. They enable down-stream tasks like data science over a large collection of documents. We extracted over 100K facts (predicates) from only 1000 ESG reports.

This work can be easily extended to include domains other than ESG. It can also be extended towards multi-modality by including text data. We leave for future exploration, the use of statements in downstream tasks like QA or document summarization.

\section*{Limitations}
\label{sec:limitations}

Although, the ideas and the techniques we describe in this paper are domain agnostic, we limit the scope of this paper to the domain of corporate Environment, Social, and Governance (ESG) reports. This choice is motivated by two observations. First, corporations report valuable quantitative data regarding their efforts to improve their carbon emissions, working conditions, and company culture in ESG reports. These reports contain valuable information regarding the environmental impact of businesses, and the urgency of climate change motivates us to target this domain. Secondly, there is a large variety and diversity of tabular representations used in these reports. Despite efforts to standardize these reports, this diversity makes the task of extracting information from these documents extremely challenging, motivating our choice. 

The scope of this work is limited to declarative, explicit knowledge only. All other kinds of knowledge such as cultural, implicit, conceptual, tacit, procedural, conditional, etc. are ignored. We focus on information which one colloquially refers to as `hard facts'. Additionally, we limit the scope of this work to quantitative statements i.e. statements whose property values are numerical quantities. We implement this restriction in the notion that we avoid qualitative statements i.e. statements which are not quantitative.

Our model training strategy was biased against large models. We trained all models for either 500K steps or 7 days using the largest possible batch size. This means smaller models learn more frequently (more epochs) than larger models. However, we do not believe this severely impacted the outcome of our experiments. Our resources were enough to recover well-known trends: improved model performance with model size and context-length.

\bibliography{main}

\appendix

\input{esg_keywords}
\onecolumn
\input{statement_examples}
\input{ted}
\input{prompt}
\input{algorithm}
\input{complex_table}
\end{document}

%% file: table_counts.tex
\begin{table}[t]
\caption{Counts of data in SemTabNet. Tasks are explained in \cref{sec:experimentalresults}.\tablefootnote{The counts differ slightly due to the manner in which the final data was harmonized. The SE Indirect 1D data consists of the 84\,890 original cells annotated from 1\,107 tables. The test/train split of tables for SE Indirect 1D was prepared by stratifying across all cell labels. This split was augmented (as described in text) to prepare data for SE Indirect 2D. The test/train split and augmentation for SE Direct was done independently.}.}
\label{tab:table_counts}
\begin{center}
\begin{small}
\begin{sc}
    \begin{tabular}{lrrr}
    \hline
        Task & Train & Test & Val \\ 
        \hline
        SE Direct & 103,455 & 11,682 & 5,445 \\ 
        SE Indirect 1D & 72,580 & 8,489 & 3,821 \\ 
        SE Indirect 2D & 93,153 & 22,839 & 4,903 \\ 
        \hline
    \end{tabular}
\end{sc}
\end{small}
\end{center}
\end{table}

%% file: table_result.tex
\begin{table*}[t]
\caption{Results of experiments performed for Statement Extraction (bold indicates the best in each experiment). The comparison between the ground truth and the model-predicted statements is encapsulated by the Tree Similarity Score ($t_s$). $t_s$ measures if two trees are similar (100\% being an exact match). For each statement, the precision, recall and F1 score (reported) of entity extraction extraction was also measured. 
For all reported values, the 99\% confidence interval, assuming a Gaussian distribution, is $\sim 0.1 \%$. The standard error of the mean in all cases is below $0.005\%$.}
\label{tab:table_result}
\begin{center}
    \begin{tabularx}{\linewidth}{clc|c|ccc|cc}
    \hline
        ~ & ~ & Context & Invalid & \multicolumn{3}{c}{Ratio Tree Edits [\%]} & \multicolumn{2}{c}{Average [\%]} 
        \\
        Task & Model & Length & Output [\%] & Insert & Delete & Rename & \multicolumn{1}{c}{F1} & \multicolumn{1}{c}{$t_{s}$}\\
        \hline
        
        ~ & Falcon-40b & 2048 & 12.59 & 45.69 & 28.77 & 25.54 & 17.94 & 0.15 \\ 
        Baseline & Llama-2-13b & 4096 & 17.93 & 79.95 & 5.78 & 14.27 & \textbf{37.94} & 5.29 \\ 
        In-Context & Llama-2-70b & 4096 & 24.82 & 89.65 & 2.56 & 7.79 & 3.18 & 6.31 \\
        1-shot & Mistral-7b & 8192 & 21.92 & 53.37 & 18.76 & 27.87 & 16.92 & 11.57 \\ 
        ~ & Mixtral-8x7b & 8192 & 19.20 & 56.39 & 18.06 & 25.56 & 6.51 & \textbf{21.07} \\ 
        \hline
        ~ & SEM-T5-small & 512 & 00.00 & 98.13 & 00.00 & 1.87 & 62.32 & 00.86 \\ 
        Indirect 1D & SEM-T5-base & 512 & 00.00 & 83.95 & 01.68 & 14.37 & 83.46 & 09.21 \\ 
        ~ & SEM-T5-large & 512 & 00.00 & 34.68 & 12.03 & 53.30 & \textbf{94.67} & \textbf{55.68} \\ 
        ~ & SEM-T5-3b & 512 & 00.00 & 36.70 & 23.24 & 40.05 & 90.49 & 22.24 \\ 
        \hline
        ~ & SEM-T5-small & 512 & 64.62 & 17.34 & 13.36 & 69.30 & 97.06 & 75.15 \\ 
        ~ & SEM-T5-base & 512 & 57.85 & 15.53 & 21.60 & 62.86 &  96.85 & 73.87 \\ 
        ~ & SEM-T5-large & 512 & 61.81 & 09.58 & 22.80 & 67.62 & 97.55 & 80.83 \\ 
        Indirect 2D & SEM-T5-3b & 512 & 50.88 & 08.00 & 28.40 & 63.59 & \textbf{97.38} & \textbf{81.76} \\ 
        ~ & SEM-T5-small & 1024 & 58.37 & 18.53 & 18.71 & 62.75 & 95.85 & 68.45 \\ 
        ~ & SEM-T5-base & 1024 & 46.39 & 17.80 & 16.04 & 66.16 & 96.15 & 69.27 \\ 
        ~ & SEM-T5-large & 1024 & 53.33 & 08.20 & 17.00 & 74.79 & 97.53 & 79.89 \\ 
        \hline
        ~ & SEM-T5-small & 512 & 00.00 & 98.14 & 00.04 & 01.82 & 60.65 & 00.62 \\ 
        ~ & SEM-T5-base & 512 & 00.00 & 97.86 & 00.06 & 02.09 & 68.62 & 04.46 \\ 
        ~ & SEM-T5-large & 512 & 00.00 & 98.18 & 00.02 & 01.80 & 67.41 & 04.23 \\ 
        Direct & SEM-T5-3b & 512 & 00.00 & 97.98 & 0.01 & 02.01 & 70.06 & 03.47 \\ 
        ~ & SEM-T5-small & 1024 & 00.00 & 92.93 & 00.14 & 06.93 & 70.35 & 02.98 \\ 
        ~ & SEM-T5-base & 1024 & 00.00 & 88.42 & 00.22 & 11.35 & \textbf{76.99} & \textbf{11.11} \\ 
        ~ & SEM-T5-large & 1024 & 00.00 & 89.34 & 00.21 & 10.45 & 76.59 & 06.06 \\ 
        \hline
    \end{tabularx}
\end{center}
\end{table*}

%% file: esg_keywords.tex
\section{ESG Keywords}
\label{appendix:esgkeywords}
\subsection*{Environment}
\input{keywords_environment}
\subsection*{Social}
\input{keywords_social}
\subsection*{Governance}
\input{keywords_governance}

%% file: keywords_environment.tex
\begin{enumerate}
\setlength\itemsep{-0.5 em}
\item \textbf{Scope 1 GHG Emissions}\\Scope 1 are all direct emissions  from the activities of an organization under their control. This includes fuel combustion on site such as gas boilers, fleet vehicles and air-conditioning leaks.
 
\item \textbf{Scope 2 GHG Emissions Market Volume}\\Scope 2 are indirect emissions from electricity purchased and used by the organization. Emissions are created during the production of the energy and eventually used by the organization. A market-based method reflects emissions from electricity that companies have actively chosen to purchase or reflects their lack of choice.
 
\item \textbf{Scope 2 GHG Emissions Location Volume}\\Scope 2 emissions are indirect emissions from the generation of purchased energy. A location-based method reflects the average emissions intensity of grids on which energy consumption occurs (using mostly grid-average emission factor data)
\item \textbf{Scope 2 GHG Emissions Other Volume}\\Scope 2 emissions are indirect emissions from the generation of purchased energy. Overall, if not clearly defined whether it is market-based calculation or location-based calculation
\item \textbf{Scope 3 GHG Emissions}\\Scope 3 emissions are all other indirect emissions (excluding Scope 2) that occur in the value chain of the reporting company, including both upstream and downstream emissions. 
\item \textbf{Environmental Restoration and Investment Initiatives Monetary Value}\\The fields represent the monetary value spent on environmental initiatives.
\item \textbf{Total Water Discharged}\\The fields represent the overall volume of water discharged by a company.
\item \textbf{Total Water Withdrawal}\\The fields represent the total volume of water withdrawn by a company.
\item \textbf{Total Water Recycled}\\The fields represent the total volume of water recycled or reused by a company.
\item \textbf{Toxic Air Emissions - NOx}\\The fields represent the total amount of nitrous oxide (NOx )emissions emitted by a company.

\item \textbf{Toxic Air Emissions - SOx}\\The fields represent the total amount of sulfur oxide (Sox) emissions emitted by a company.

\item \textbf{Toxic Air Emissions - Overall}\\The fields represent the total amount of air emissions emitted by a company.
\item \textbf{Toxic Air Emissions - VOC}\\The fields represent the total amount of volatile organic compound (VOC) emissions emitted by the company.
\item \textbf{Hazardous Waste - Disposed to Aquatic}\\The fields represent the total amount of hazardous waste disposed to aquatic environment.
\item \textbf{Hazardous Waste - Disposed to Land }\\The fields represent the total amount of hazardous waste disposed to non aquatic or land environment.
\item \textbf{Hazardous Waste - Total Recycled}\\The fields represent the total amount of hazardous waste recycled.
\item \textbf{Hazardous Waste - Total Amount Generated}\\The fields represent the total amount of hazardous waste generated by a company.
\item \textbf{Hazardous Waste - Total Amount Disposed}\\The fields represent the total amount of hazardous waste disposed.
\item \textbf{Non-Hazardous Waste - Disposed to Aquatic}\\The fields represent the total amount of non-hazardous waste disposed to the aquatic environment.
\item \textbf{Non-Hazardous Waste - Disposed to Land }\\The fields represent the total amount of non-hazardous waste to non aquatic or land environment
\item \textbf{Non-Hazardous Waste - Total Recycled}\\The field represents the total amount of non-hazardous waste recycled.
\item \textbf{Non-Hazardous Waste - Total Amount Generated}\\The fields represent the total amount of non-hazardous waste Generated by a company.
\item \textbf{Non-Hazardous Waste - Total Amount Disposed}\\The fields represent the total amount of non-hazardous waste disposed.
\item \textbf{Total Waste Produced}\\The fields represent the total amount of waste produced by a company.
\item \textbf{Total Waste Recycled}\\The fields represents the total  amount of waste recycled by a company.
\item \textbf{Total Waste Disposed}\\This fields represent the total amount of waste disposed by a company.
\item \textbf{Number of Sites in Water Stress Areas}\\The field represents the number of sites located in water stress areas.
\item \textbf{E-Waste Produced }\\The field identifies the mass volume of f E- waste produced which are electronic products that are unwanted, not working, and nearing or at the end of their life. Examples of electronic waste include, but not limited to : computers, printers, monitors, and mobile phones
\item \textbf{E-Waste Recycled}\\The field identifies the mass volume of E- Waste Recycled.
\item \textbf{E-Waste Disposed}\\The field identifies the mass volume of E- waste disposed.
\item \textbf{Number of Sites Operating in Protected and/or High Biodiversity Areas}\\The field identifies the number of sites or facilities owned,leased, managed in or adjacent  to protected areas and areas of high biodiversity value outside protected areas.
\item \textbf{Impacted Number of Species on International Union of Conservation of Nature (IUCN) List }\\The field identifies the number of impacted species on International Union of Conservation of Nature (IUCN) red list. 
\item \textbf{Impacted Number of Species on National listed Species}\\The field identifies the number of impacted species on National Listed Species.
\item \textbf{Baseline Level}\\The field identifies the value at baseline or year that target is set against.
\item \textbf{Target Year}\\The field identifies the year in which the renewable energy goal is set to be completed.
\item \textbf{Target Goal}\\The field identifies the target goal for renewable energy.
\item \textbf{Actual Achieved}\\The fields identifies the actual value achieved for the renewable energy goal.
\item \textbf{Baseline Level}\\The field identifies the baseline emissions value.
\item \textbf{Target Year}\\The field identifies the  year in which GHG emission goal is set to be completed.
\item \textbf{Target Goal}\\The field identifies the target goal for GHG emission reduction.
\item \textbf{Actual Achieved}\\The field identifies the value achieved of GHG emissions reduced compare - in metric tons.
\end{enumerate}

%% file: keywords_social.tex
\begin{enumerate}
\setlength\itemsep{-0.5 em}
\item \textbf{Training Hours Per Employee}\\The fields identifies the numerical value of training hours per employee.
\item \textbf{Training Hours Annually}\\The fields identifies the numerical values of training hours conducted within a year.
\item \textbf{Lost Time Injury Overall Rate}\\The fields identifies the total number of injuries that caused the employees and contractors to lose at least a working day.
\item \textbf{Lost Time Injury Rate Contractors}\\The fields identifies the number of injuries that caused the contractors to lose at least a working day. 
\item \textbf{Lost Time Injury Rate Employees}\\The fields identifies the number of injuries that caused the employees to lose at least a working day.
\item \textbf{Employee Fatalities}\\The fields identifies the number of employee fatalities during a one year period.
\item \textbf{Contractor Fatalities}\\The fields identifies the number of contractor fatalities during a one year period.
\item \textbf{Public Fatalities}\\The fields identifies the number of general public fatalities during a one year period.
\item \textbf{Number of Other Fatalities}\\The fields identifies the number of fatalities during a one year period not broken down by employee, contractor, or public.
\item \textbf{Total Incident Rate Overall Workers}\\The field identifies the number of work-related injuries per 100 overall workers during a one year period for both employees and contractors.
\item \textbf{Total Incident Rate Contractors}\\The field identifies the number of contractor work-related injuries per 100 overall workers during a one year period.
\item \textbf{Total Incident Rate Employees}\\The field identifies the number of work-related injuries per 100 overall workers during a one year period for employees.
\item \textbf{Employee Turnover - Gender Male Rate}\\The field identifies the absolute number turnover rate by males in a company .
\item \textbf{Employee Turnover - Gender Female Rate}\\The field identifies the absolute number turnover rate by females in a company.
\item \textbf{Employee Turnover Overall Rate}\\The field identifies the absolute number turnover rate for overall  employees in a company.
\item \textbf{Median Gender Pay Gap - Global}\\The field identifies the gender pay gap median value of the company at a global level.
\item \textbf{Mean Gender Pay Gap - Global}\\The field identifies the gender pay gap mean or average value of the company at a global level.
\item \textbf{Median Gender Pay Gap by Location}\\The field represents the gender pay gap median value of the company at a location or country level.
\item \textbf{Mean Gender Pay Gap by Location}\\The field represents the gender pay gap mean/average value of the company at a location or country level.
\item \textbf{Employee Turnover by Age - Lower Value}\\The field Identifies the minimum age in a given range for employee turnover statistics.
\item \textbf{Employee Turnover by Age - Upper Value}\\The field identifies the maximum age in a given range for employee turnover statistics.
\item \textbf{Employee Turnover by Age - Rate}\\The field identifies the employee turnover rate.
\item \textbf{Employee Turnover by Location Rate}\\The field identifies the absolute number of employee turnover rate by location.
\item \textbf{Workforce Breakdown Rate}\\The field identifies the absolute number of employees of a company based on seniority, ethnicity or gender.
\item \textbf{Workforce Breakdown Job Category Data: Value (ABS)}\\The field represents the employee count absolute value at a category level within a workforce.
\item \textbf{Number Of Product Recalls}\\The fields identifies the number of product recalls.
\item \textbf{Product Recalls Annual Recall Rate}\\The fields identifies the product recall rate of a company.
\end{enumerate}

%% file: keywords_governance.tex
\begin{enumerate}
\setlength\itemsep{-0.5 em}
\item \textbf{Percentage of Negative Votes on Pay Practices Year}\\
\item \textbf{Board of Director Term Limit}\\The field identifies maximum amount of years  a board member can serve.
\item \textbf{Board of Director Term Duration}\\The field identifies number of years a board member can serve before reelection. 
\item \textbf{Auditor Election Year}\\The field identifies when the current lead auditor elected.
\item \textbf{Independent Auditor Start Year}\\The field represents the start year the company started having the audit company as its independent auditor.
\item \textbf{Average/Mean Compensation of Company Employees-Global}\\The field represents the average or mean compensation for company employeesat a global level.
\item \textbf{Ratio Average Compensation of CEO to Employee - CEO- Global}\\The field represents the ratio between the compensation paid to the companies CEO and the average compensations received by employees at a global level.
\item \textbf{Compensation of Company Employees by Location}\\The field identifies the average compensation for company employees at a location level.
\item \textbf{Number of Suppliers Complying with Code of Conduct}\\The field identifies the number of suppliers that comply with companies supplier code of conduct.
\item \textbf{Share Class Numeric}\\The field identifies the share class numeric component.
\item \textbf{Voting Rights}\\The field identifies the  number of voting rights per each share of stock within each class.
\item \textbf{Shares Outstanding}\\The field identifies the number of shares outstanding within a companies common stock.
\item \textbf{Chairman Effective Begin Year}\\The field indicates the year when the current chairman assume his or her position. This field is used if a full effective date is not available.
\item \textbf{Chairman Effective End Year}\\The field indicates the year when the chairman left the position.
\item \textbf{CEO Effective Begin Year}\\The field identifies the year the CEO assumed his or her position.
\item \textbf{CEO Effective End Year}\\The field indicates the year when the CEO left the position.
\item \textbf{CEO Compensation Salary}\\The field identifies the current CEO salary.
\item \textbf{CEO Compensation Overall}\\The field identifies the CEO's overall compensation including salary, bonuses and all awards.
\item \textbf{CEO Cash Bonus}\\The field identifies the  cash bonus value for the CEO.
\item \textbf{CEO Stock Award Bonus}\\The CEO Stock Award Bonus value
\item \textbf{CEO Option Awards}\\The  CEO Option Awards bonus value
\item \textbf{CEO Other Awards}\\The fields identifies other compensation outside of salary, cash bonus, stock award bonus and option awards. This could include change in pension and values categorized as "all other compensation"
\item \textbf{CEO Pension}\\The fields identifies the CEO pension amount.
\item \textbf{Cash Severance Value}\\The fields identifies the amount of cash the severance policy for each category.
\item \textbf{Total Severance Value}\\The fields identifies the total value amount of the severance policy.
\item \textbf{CEO Share Ownership}\\The field identifies the number of shares the CEO owns in the company.
\item \textbf{CEO Share Class Numeric}\\The field identifies the share class numeric component.
\item \textbf{Board Member Age}\\The field identifies the age of the members of the board.
\item \textbf{Board Member Term in Years}\\The fields identifies how long the individual board member has been on the board which is determined in years.
\item \textbf{Board Member Effective Year (Director Since)}\\The fields identifies the year the individual board member started serving on the board.
\item \textbf{Board Profile As of Year}\\The field identifies the year of the board information. An example would be the year of the proxy statement.
\item \textbf{Participation On Other Company Board}\\The field identifies the number of boards a member is part of outside of the organization.
\item \textbf{For Value Negative Votes on Directors}\\The field identifies the number of for value votes the director received.
\item \textbf{Against Value Negative Votes on Directors}\\The  field identifies the number of against votes the director received.
\item \textbf{Abstain Value Negative Votes on Directors}\\The field identifies the number of votes that were abstained for a given director.
\item \textbf{Broker Non Vote Value Negative Votes on Directors}\\The field identifies the number of broker non votes for given director.
\item \textbf{Number of Board Meetings Attended by Board Member}\\The field identifies the number of board meetings attended by a board member.
\item \textbf{Number of Board Meetings Held by Company}\\The field identifies the number of board meetings held by a company while member was on the board.
\item \textbf{Total Members on Board per Skill Set}\\The field identifies the number of board members within a specific skillset type.
\end{enumerate}

%% file: statement_examples.tex
\section{Examples of Statements}
\label{appendix:examplestatement}

A \statement/ is complete when it contains all the predicates needed to completely specify objective knowledge pertaining to a subject, i.e. a \statement/ includes all co-dependent predicates. We borrow this notion of completeness from the fields of natural science. An important implication of these definitions is that within a single statement, multiple predicates cannot carry information about the same `property'. This implies, for example, multiple measurements of the same variable in \textit{n} different conditions will lead to \textit{n} different statements. While complete \statements/ are extremely valuable, we find that incomplete \statements/ are quite resourceful, especially as we apply our ideas to domains outside of natural science.

Examples of statements from other domains are shown below.

\textit{Basic Sciences:} Consider the following piece of text or unstructured data. ``At a pressure of one atmosphere (atm), water freezes (solidifies) at 0 $^{\circ}$C and water boils at 100 $^{\circ}$C.'' We note that to completely describe the phase changes of water, we need to specify both temperature and pressure. Leaving any one of temperature or pressure out makes the information regarding phase change incomplete. This information is presented as \statements/ in the Tables \cref{tab:statment_water_freezing} and \cref{tab:statment_water_boiling}. This example demonstrates that multiple statements can be extracted from even single sentences. 

\begin{table*}[!h]
\begin{center}
    \caption{Example Statement from Material Science: Phase change of water from solid to liquid.}
    \begin{tabular}{ccccc}
    \hline
    Subject & Subject Value & Property  & Property Value & Unit\\
    \hline
    Chemical & Water & freezing temperature & 0 & $^{\circ}$C \\
    Chemical & Water & pressure & 1 &  atmosphere \\
    \hline
    \end{tabular}
    \label{tab:statment_water_freezing}
    \end{center}
\end{table*}

\begin{table*}[!h]
\begin{center}
    \caption{Example Statement from Material Science: Phase change of water from liquid to gas.}
    \begin{tabular}{ccccc}
    \hline
    Subject & Subject Value & Property  & Property Value & Unit\\
    \hline
    Chemical & Water & boiling temperature & 100 & $^{\circ}$C \\
    Chemical & Water & pressure & 1 &  atmosphere \\
    \hline
    \end{tabular}
    \label{tab:statment_water_boiling}
    \end{center}
\end{table*}

\textit{Physics: } 

\begin{table*}[!h]
\begin{center}
    \caption{Example Statement from Physics: Speed of light.}
    \begin{tabular}{ccccc}
    \hline
    Subject & Subject Value & Property  & Property Value & Unit\\
    \hline
    Boson & Light & speed & 299\,792\,458 & $m s^{-1}$ \\
    Boson & Light & medium & vacuum &   \\
    \hline
    \end{tabular}
    \end{center}
\end{table*}

Independent properties make independent statements, as shown below.

\begin{table*}[!h]
\begin{center}
    \caption{Example Statement from Physics: Mass of electron.}
    \begin{tabular}{ccccc}
    \hline
    Subject & Subject Value & Property  & Property Value & Unit\\
    \hline
    Fermion & Electron & Mass & $9.1093837015 \times 10^{-31}$ & kg \\
    \hline
    \end{tabular}
    \end{center}
\end{table*}

\begin{table*}[!h]
\begin{center}
    \caption{Example Statement from Physics: Charge of electron.}
    \begin{tabular}{ccccc}
    \hline
    Subject & Subject Value & Property  & Property Value & Unit\\
    \hline
    Fermion & Electron & Electric Charge & $ - 1.602176634 \times 10^{-19}$ & C \\
    \hline
    \end{tabular}
    \end{center}
\end{table*}

\begin{table*}[!h]
\begin{center}
    \caption{Example Statement from Physics: Charge of electron.}
    \begin{tabular}{ccccc}
    \hline
    Subject & Subject Value & Property  & Property Value & Unit\\
    \hline
    Fermion & Electron & Spin & 1/2 & $h$ \\
    \hline
    \end{tabular}
    \end{center}
\end{table*}

%% file: ted.tex
\section{TED Similarity Score}
\label{appendix:ted}
\subsection{Creating Trees}

The \statement/ data structure can be viewed in many representations: hypergraphs, tree, table, records, and transforming the representation of this data structure in other formats is trivial.

In our setup, when represented as a tree, all nodes in a \statement/ has four attributes: \textit{name}, \textit{type}, \textit{value}, and \textit{parent}. We start a tree with the root node with name as `/root', type as `root', and no value. This node does not have any parent node. Next, the statement nodes emerge as branches from the root. Each statement node has a name like `/root/s0' or `/root/s2' (here, `s' indicates that this is a statement node and the number acts as an index), type as `statement', no value and the root node as its parent.  Further, attached to each statement node are predicate node(s) with names like `/root/s1/p0' or `/root/s0/p3', type as `predicate', no value and a statement node as its parent. Finally, in our current implementation, each predicate node has five children nodes attached to it. These leaf nodes can be of type: subject, subject-value, property, property-value, unit and the value attribute is populated with the actual value. The leaf nodes may have names like `/root/s2/p1/subject' or `/root/s0/p3/property-value'. In this representation, the name of a node completely determines the location of the node in a tree.

As an example, we show the tree structure for the statements shown in \cref{fig:text_table_to_statements}:

\begin{mdframed}[backgroundcolor=gray!10]
\small
\begin{verbatim}
Node('/root', type='root', value=None)
|-- Node('/root/s0', type='statement', value=None)
|   |-- Node('/root/s0/p0', type='predicate', value=None)
|   |   |-- Node('/root/s0/p0/Subject', type='Subject', value='Organization')
|   |   |-- Node('/root/s0/p0/Subject Value', type='Subject Value', value='XYZ')
|   |   |-- Node('/root/s0/p0/Property', type='Property', value='scope 1 emissions')
|   |   |-- Node('/root/s0/p0/Property Value', type='Property Value', value='3.3')
|   |   |-- Node('/root/s0/p0/Unit', type='Unit', value='million metric tons of CO2e')
|   |-- Node('/root/s0/p1', type='predicate', value=None)
|       |-- Node('/root/s0/p1/Subject', type='Subject', value='Organization')
|       |-- Node('/root/s0/p1/Subject Value', type='Subject Value', value='XYZ')
|       |-- Node('/root/s0/p1/Property', type='Property', value='time')
|       |-- Node('/root/s0/p1/Property Value', type='Property Value', value='2020')
|       |-- Node('/root/s0/p1/Unit', type='Unit', value='year')
|-- Node('/root/s1', type='statement', value=None)
    |-- Node('/root/s1/p0', type='predicate', value=None)
    |   |-- Node('/root/s1/p0/Subject', type='Subject', value='Organization')
    |   |-- Node('/root/s1/p0/Subject Value', type='Subject Value', value='XYZ')
    |   |-- Node('/root/s1/p0/Property', type='Property', value='scope 1 emissions')
    |   |-- Node('/root/s1/p0/Property Value', type='Property Value', value='2.5')
    |   |-- Node('/root/s1/p0/Unit', type='Unit', value='million metric tons of CO2e')
    |-- Node('/root/s1/p1', type='predicate', value=None)
        |-- Node('/root/s1/p1/Subject', type='Subject', value='Organization')
        |-- Node('/root/s1/p1/Subject Value', type='Subject Value', value='XYZ')
        |-- Node('/root/s1/p1/Property', type='Property', value='time')
        |-- Node('/root/s1/p1/Property Value', type='Property Value', value='2021')
        |-- Node('/root/s1/p1/Unit', type='Unit', value='year')
\end{verbatim}
\end{mdframed}

\subsection{Computing Tree Similarity Score}
\label{appendix:treesimilarityscore}

For comparing two statement trees, we setup strict costs for each edit operation. The predictions are maximally punished for any structural deviation from the ground truth, i.e. deletion and insertion each have a cost of 1. For renaming of the node's value attribute, we only allow two nodes to be renamed if they are of the same type. If both nodes' value attribute is of type string, then we calculate a normalized Levenshtein edit distance between the two strings.

If both nodes' value attribute is of numerical type, then the two values are directly compared. In this case, the cost is 0 if the two values are the same, and 1 in all other cases. If the value attribute of both the ground truth and the prediction node is empty, then the cost operation is also 0. We denote TED with $t$. We define normalized TED (nTED or $\overline{t}$) as the ratio of the distance to the number of edits between two trees. Using the normalized TED, a normalized Tree Similarity score can be computed as $t_{s} = 1 - \overline{t}$.

Consider comparing the trees for the two statements $s0$ and $s1$, from the example above. These two trees differ only in their numeric value but are otherwise similar to each other. Two edits are required to convert one tree into another: one corresponding to the property-value of `time' and the other corresponding to the property-value of `scope 1 emissions'. If the numeric values are interpreted as floats, then our strict setup will maximally punish for each edit giving an edit distance of 2 renaming, 0 deletions, and 0 insertions. The normalized tree edit distance (ratio of distance to total number of edits) would be 2 / 2 = 1. Thus, the TED similarity score would be 1 - 1 = 0. 

However, our model outputs numeric values as strings, which can be compared via normalized Levenshtein distance. Then, the first rename edit of year values will give a distance of 1/4 = 0.25, and the other rename edit will give a distance of 2/3 = 0.66. In this case, the total tree edit distance is 0.9166, the normalized tree edit distance is 0.4583. This gives a TED similarity score of 0.54. We will interpret this by saying that ``the two tree (when the numeric value are interpreted as strings) are 54\% similar to each other''. Given that the two trees are similar in their structure and only differ in their numeric values, this shows that our setup of TED similarity score is very strict. 

For illustrative purposes, let us consider another example. We consider that the $s0$ in the above example is the ground truth statement:

\begin{mdframed}[backgroundcolor=blue!10]
\small
\begin{verbatim}
Node('/root', type='root', value=None)
|-- Node('/root/s0', type='statement', value=None)
|   |-- Node('/root/s0/p0', type='predicate', value=None)
|   |   |-- Node('/root/s0/p0/subject', type='subject', value='Organization')
|   |   |-- Node('/root/s0/p0/subject_value', type='subject_value', value='XYZ')
|   |   |-- Node('/root/s0/p0/property', type='property', value='scope 1 emissions')
|   |   |-- Node('/root/s0/p0/property_value', type='property_value', value='3.3')
|   |   |-- Node('/root/s0/p0/unit', type='unit', value='million metric tons of CO2e')
|   |-- Node('/root/s0/p1', type='predicate', value=None)
|       |-- Node('/root/s0/p1/subject', type='subject', value='Organization')
|       |-- Node('/root/s0/p1/subject_value', type='subject_value', value='XYZ')
|       |-- Node('/root/s0/p1/property', type='property', value='time')
|       |-- Node('/root/s0/p1/property_value', type='property_value', value='2020')
|       |-- Node('/root/s0/p1/unit', type='unit', value='year')
\end{verbatim}
\end{mdframed}

And we have a model which makes the following prediction:

\begin{mdframed}[backgroundcolor=orange!10]
\small
\begin{verbatim}
Node('/root', type='root', value=None)
|-- Node('/root/s1', type='statement', value=None)
    |-- Node('/root/s1/p0', type='predicate', value=None)
        |-- Node('/root/s1/p0/subject', type='subject', value='Organization')
        |-- Node('/root/s1/p0/subject_value', type='subject_value', value='XYZ')
        |-- Node('/root/s1/p0/property', type='property', value='scope 2 emissions')
        |-- Node('/root/s1/p0/property_value', type='property_value', value='3.3')
        |-- Node('/root/s1/p0/unit', type='unit', value='million metric tons of CO2e')
\end{verbatim}
\end{mdframed}

We observe that the predicted tree is missing an entire predicate with time property. This happens when models stop generating new tokens. Compared to the previous example, the ground truth and model prediction have a major structural deviation. In addition, the model also made a mistake in the value of the `property' node. Instead of `scope 1 emissions' as in ground truth, the model predicted `scope 2 emissions'. 

To convert one tree into another, we need a total of 7 edits: six nodes need to be deleted (or inserted) (5 leaf nodes and 1 predicate node) and 1 renaming edit. All deletions or insertions have equal score of 1 each, and the renaming costs 1/17 $\approx$ 0.0588. The total tree edit distance becomes 6.0588, the normalized tree edit distance is 0.8655. This gives us a tree similarity score of 0.1344. We interpret that the two trees are only 13\% similar to each other.

%% file: prompt.tex
\section{Baseline Experiments}
\label{appendix:prompt}

\textbf{Example of successful statement extraction:}

\begin{table*}[!h]
\center
\caption{Table with simple layout from page 68 of the 2022 ESG report from Splunk Inc. }
\begin{tabular}{|l|r|r|}
\hline
0 & 1 & 2 \\
\hline
Emissions Scope  & FY21  & FY22  \\
Scope 1 Direct Emissions  & 24  & 374  \\
Scope 2 Indirect Emissions  & 3,686  & 3,257  \\
Scope 3 Other Indirect Emissions  & 11,430  & 7,938  \\
Total  & 15,140  & 11,569  \\
\hline
\end{tabular}
\end{table*}

Consider the above table, with a simple layout, from the 2022 ESG report of Splunk Inc.

We prompt Mixtral with the above table using the following prompt. For rendering, we replace our line-break token `<br>' with actual line-breaks and remove some aspect of the example statement for brevity.

\begin{mdframed}[backgroundcolor=gray!10]
\tiny
\begin{verbatim}
<table>
| 0                                              |        1 |        2 |        3 |
|------------------------------------------------|----------|----------|----------|
| Public Metrics                                 | 2019     | 2020     | 2021     |
| Public Fatalities Due to Electrical Contacts   |    4     |    2     |    5     |
| Employee Metrics                               | 2019     | 2020     | 2021     |
| Employee DART Rate                             |    0.412 |    0.31  |    0.43  |
| Total Employee Recordable Incident Rate (TRIR) |    0.673 |    0.576 |    0.648 |
| Employee Lost Time Incident Rate (LTIR)        |    0.329 |    0.245 |    0.333 |
| Employee Severity Rate                         |   22.719 |   15.284 |   19.113 |
| Total Employee OSHA Recordable Events          |  129     |  106     |  113     |
| Employee Fatalities                            |    1     |    0     |    0     |
</table>
<response>
...
| property                                                   |   property_value | unit   | subject   | subject_value   |
|------------------------------------------------------------|------------------|--------|-----------|-----------------|
| Employee Metrics : Employee Lost Time Incident Rate (LTIR) |            0.329 |        |           |                 |
| time                                                       |         2019     |        |           |                 |
<sep>
| property                                                   |   property_value | unit   | subject   | subject_value   |
|------------------------------------------------------------|------------------|--------|-----------|-----------------|
| Employee Metrics : Employee Lost Time Incident Rate (LTIR) |            0.245 |        |           |                 |
| time                                                       |         2020     |        |           |                 |
<sep>
| property                                                   |   property_value | unit   | subject   | subject_value   |
|------------------------------------------------------------|------------------|--------|-----------|-----------------|
| Employee Metrics : Employee Lost Time Incident Rate (LTIR) |            0.333 |        |           |                 |
| time                                                       |         2021     |        |           |                 |
<sep>
| property                                  |   property_value | unit   | subject   | subject_value   |
|-------------------------------------------|------------------|--------|-----------|-----------------|
| Employee Metrics : Employee Severity Rate |           22.719 |        |           |                 |
| time                                      |         2019     |        |           |                 |
<sep>
| property                                  |   property_value | unit   | subject   | subject_value   |
|-------------------------------------------|------------------|--------|-----------|-----------------|
| Employee Metrics : Employee Severity Rate |           15.284 |        |           |                 |
| time                                      |         2020     |        |           |                 |
<sep>
| property                                  |   property_value | unit   | subject   | subject_value   |
|-------------------------------------------|------------------|--------|-----------|-----------------|
| Employee Metrics : Employee Severity Rate |           19.113 |        |           |                 |
| time                                      |         2021     |        |           |                 |
...
</response>

<table>
| 0                                | 1      | 2      |
|----------------------------------|--------|--------|
| Emissions Scope                  | FY21   | FY22   |
| Scope 1 Direct Emissions         | 24     | 374    |
| Scope 2 Indirect Emissions       | 3,686  | 3,257  |
| Scope 3 Other Indirect Emissions | 11,430 | 7,938  |
| Total                            | 15,140 | 11,569 |
</table>
<response>

\end{verbatim}
\end{mdframed}

\newpage
The model output for the above prompt with greedy decoding was:

\begin{mdframed}[backgroundcolor=orange!10]
\tiny
\begin{verbatim}
| property                                                  |   property_value | unit   | subject   | subject_value   |
|-----------------------------------------------------------|------------------|--------|-----------|-----------------|
| Emissions Scope : Scope 1 Direct Emissions                |               24 |        |           |                 |
| time                                                      |             FY21 |        |           |                 |
<sep>
| property                                                  |   property_value | unit   | subject   | subject_value   |
|-----------------------------------------------------------|------------------|--------|-----------|-----------------|
| Emissions Scope : Scope 1 Direct Emissions                |              374 |        |           |                 |
| time                                                      |             FY22 |        |           |                 |
<sep>
| property                                                  |   property_value | unit   | subject   | subject_value   |
|-----------------------------------------------------------|------------------|--------|-----------|-----------------|
| Emissions Scope : Scope 2 Indirect Emissions              |            3,686 |        |           |                 |
| time                                                      |             FY21 |        |           |                 |
<sep>
| property                                                  |   property_value | unit   | subject   | subject_value   |
|-----------------------------------------------------------|------------------|--------|-----------|-----------------|
| Emissions Scope : Scope 2 Indirect Emissions              |            3,257 |        |           |                 |
| time                                                      |             FY22 |        |           |                 |
<sep>
| property                                                  |   property_value | unit   | subject   | subject_value   |
|-----------------------------------------------------------|------------------|--------|-----------|-----------------|
| Emissions Scope : Scope 3 Other Indirect Emissions        |           11,430 |        |           |                 |
| time                                                      |             FY21 |        |           |                 |
<sep>
| property                                                  |   property_value | unit   | subject   | subject_value   |
|-----------------------------------------------------------|------------------|--------|-----------|-----------------|
| Emissions Scope : Scope 3 Other Indirect Emissions        |            7,938 |        |           |                 |
| time                                                      |             FY22 |        |           |                 |
<sep>
| property                                                  |   property_value | unit   | subject   | subject_value   |
|-----------------------------------------------------------|------------------|--------|-----------|-----------------|
| Emissions Scope : Total                                   |           15,140 |        |           |                 |
| time                                                      |             FY21 |        |           |                 |
<sep>
| property                                                  |   property_value | unit   | subject   | subject_value   |
|-----------------------------------------------------------|------------------|--------|-----------|-----------------|
| Emissions Scope : Total                                   |           11,569 |        |           |                 |
| time                                                      |             FY22 |        |           |                 |
</response>
\end{verbatim}
\end{mdframed}

This is an example of correct statement extraction. For the same table with a different example in the prompt, the output of the same model was:

\begin{mdframed}[backgroundcolor=orange!10]
\tiny
\begin{verbatim}
| property                                                                                                                                                                                                                              
\end{verbatim}
\end{mdframed}
This is an invalid output without any correct markdown structure or content. This shows that the in-context approach is sensitive to the prompt and thus is not robust.

%% file: algorithm.tex
\section{Algorithm for Statement Extraction}
\label{appndix:tablecellannotation}

We present the algorithm we used to extract statements. For this algorithm, the inputs are the original table and the labels table.

\begin{algorithm}
\small
\begin{algorithmic}[1]
\Procedure{Extract Statements}{Table, LabelsTable}
\State \textbf{Input:} Table, LabelsTable: Table and Table of cell annotations
\State AllStatements $\gets$ empty list
\ForAll{row in LabelsTable}
\ForAll{column in LabelsTable}
\If{LabelsTable[row][column] = Property Value}
\State \textbf{Search in the same row and column} for (Sub)-Property
\If{Property is found}
\State \textbf{Append Headers} in hierarchy to Property, if any, starting from the minimum level
\State \textbf{Construct Statement} with Property, Row and Column
\ElsIf{SubProperty is found}
\State \textbf{Append Property} to the SubProperty
\State \textbf{Append Headers} in hierarchy to SubProperty, if any, starting from the maximum level
\State \textbf{Construct Statement} with SubProperty, Row and Column
\Else
\State Property is not found, continue to the next iteration
\EndIf
\State Append Statement to AllStatements
\EndIf
\EndFor
\EndFor
\State \textbf{Return} AllStatements
\EndProcedure
\label{alg:statement_extraction}
\caption{Extract Statements}
\end{algorithmic}
\end{algorithm}

\begin{algorithm}
\small
\begin{algorithmic}[1]
\Procedure{Construct Statement}{Row, Column, Property}
\State \textbf{Input:} Row, Column, Property: Row and Column of the Property Value, with its related Property
\State \textbf{Output:} Statement: list
\State Statement $\gets$ empty list
\State Predicate $\gets$ empty dictionary
\State Predicate [Property Value] $\gets$ \text{Table}[Row][Column]
\State Predicate [Property] $\gets$ Property
\State \textbf{Search in the same row and column}(Unit Value)
\State Predicate[Unit] $\gets$ Table[row$\sb{uv}$][column$\sb{uv}$]
\State \textbf{Search for a } Subject - Subject Value \textbf{ pair}
\State Predicate[Subject] $\gets$ Table[row$\sb{s}$][column$\sb{s}$]
\State Predicate[Subject\_Value] $\gets$ Table[row$\sb{sv}$][column$\sb{sv}$]
\State Add Predicate to the Statement
\State \textbf{Search in the same row and column}(Time Value)
\If{Time Value is found}
\State Predicate $\gets$ empty dictionary
\State Predicate [Property Value] $\gets$ \text{Table}[row$\sb{tv}$][column$\sb{tv}$]
\State Predicate [Property] $\gets$ "Time"
\State Add Predicate to the Statement
\EndIf
\State \textbf{Search for all} Key - Key Value \textbf{pairs}
\ForAll{Key - Key Value pairs found}
\State Predicate $\gets$ empty dictionary
\State Predicate[Property] $\gets$ Table[row$\sb{k}$][column$\sb{k}$]
\State Predicate[Property Value] $\gets$ Table[row$\sb{kv}$][column$\sb{kv}$]
\State Add Predicate to the Statement
\EndFor
\State \textbf{Return} Statement
\EndProcedure
\end{algorithmic}
\end{algorithm}

\begin{algorithm}
\small
\begin{algorithmic}[1]

\Procedure{Append Headers}{Row, Column, Propery, Level}
\State \textbf{Input:} Row, Column, Property, Level: Row, Column, value of a Property cell and the level of the header to search for.
\State \textbf{Output:} Property: string
\ForAll{Row$\sb{a}$ above Row}
\ForAll{Column$\sb{l}$ on the left of Column}
\If{LabelsTable[Row$\sb{a}$][Column$\sb{l}$] is a header with a higher level than Level}
\State Append Table[Row$\sb{a}$][Column$\sb{l}$] on top of Property
\If{the level of LabelsTable[Row$\sb{a}$][Column$\sb{l}$] is maximum}
\State \textbf{Return} Property
\Else
\State \textbf{Append Headers} in hierarchy to Property starting from the level of LabelsTable[Row$\sb{a}$][Column$\sb{l}$]
\State \textbf{Return} Property
\EndIf
\EndIf
\EndFor
\EndFor
\State \textbf{Return} Property
\EndProcedure
\caption{Utility function for appending section header.}
\end{algorithmic}
\end{algorithm}

\begin{algorithm}
\small
\begin{algorithmic}[1]
\Procedure{Append Property}{Row, Column, SubProperty}
\State \textbf{Input:} Row, Column, SubProperty: Row,Column and Value of a SubProperty cell
\State \textbf{Output:} Subproperty: string
\ForAll{Row$\sb{a}$ above Row}
\ForAll{Column$\sb{l}$ on the left of Column}
\If{LabelsTable[Row$\sb{a}$][Column$\sb{l}$] is a Property}
\State Append Table[Row$\sb{a}$][Column$\sb{l}$] on top of SubProperty
\State \textbf{Return} SubProperty
\EndIf
\EndFor
\EndFor
\State \textbf{Return} SubProperty
\EndProcedure
\caption{Utility function for appending property name to sub-property}
\end{algorithmic}
\end{algorithm}

\begin{algorithm}
\small
\begin{algorithmic}[1]

\Procedure{Search in the same row and column}{Row, Column, Key}
\State \textbf{Input:} Row, Column, Key: Row and Column where to search the specified Key
\State \textbf{Output:} Row$\sb{k}$, Column$\sb{k}$: Row and column of the designated Key, if found
\ForAll{Cell respectively on the Left, Above, and Right to the cell at LabelsTable[Row][Column]}
\If{Cell is Key}
\State \textbf{Return} Row, Column of Cell
\EndIf
\EndFor
\State \textbf{Return} Null
\EndProcedure
\caption{Utility function to search for related predicates}
\end{algorithmic}
\end{algorithm}

\begin{algorithm}
\small
\begin{algorithmic}[1]

\Procedure{Search for a pair}{Row, Column, Key, Key Value}
\State \textbf{Input:} Row, Column, Key: Row and Column where to search the specified Key
\State \textbf{Output:} Row$\sb{k}$, Column$\sb{k}$: Row and column of the designated Key, if found
\ForAll{Cell$\sb{kv}$ respectively on the Left, Above, and Right to the cell at LabelsTable[Row][Column]}
\If{Cell$\sb{kv}$ is Key Value}
\ForAll{Cell$\sb{k}$ in the Orthogonal Direction with respect to Cell$\sb{kv}$ from LabelsTable[Row][Column]}
\If{Cell$\sb{k}$ is Key}
\State \textbf{Return} Coordinates of Cell$\sb{k}$, Cell$\sb{kv}$
\EndIf
\EndFor
\EndIf
\EndFor
\State \textbf{Return} Null
\EndProcedure
\caption{Utility function for searching corresponding key-value.}
\end{algorithmic}
\end{algorithm}

%% file: complex_table.tex
\begin{figure}[t!]
\includegraphics[width=\linewidth]{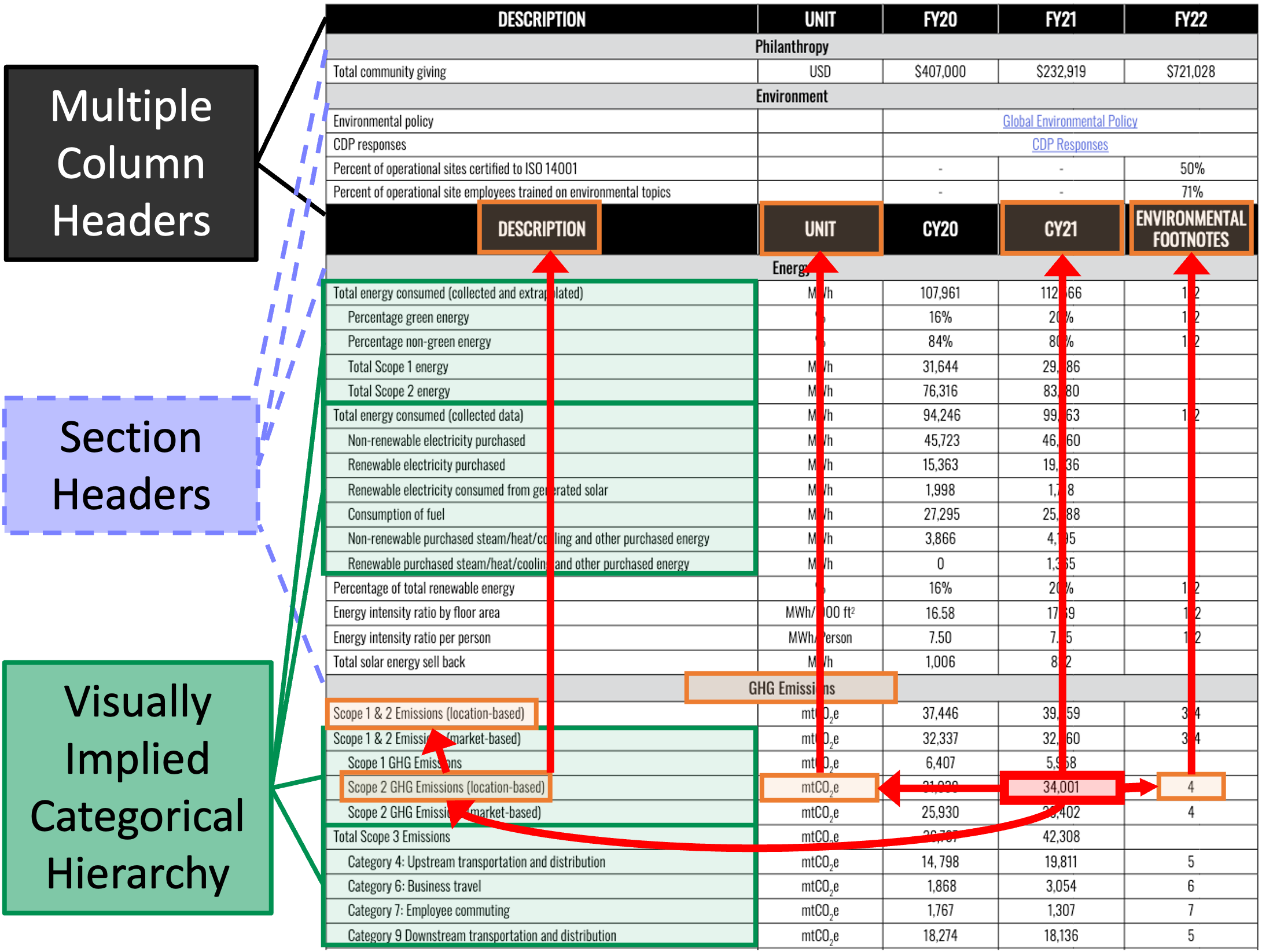}
\caption{Example table from an ESG report with a complicated layout. To extract the information content of a single cell (highlighted in red), the content and relationships (lines drawn in red) to many other cells (highlighted in orange) also needs to be understood.}
\label{fig:complex_table}
\end{figure}

%% file: main.bbl
\begin{thebibliography}{33}
\expandafter\ifx\csname natexlab\endcsname\relax\def\natexlab#1{#1}\fi

\bibitem[{Almazrouei et~al.(2023)Almazrouei, Alobeidli, Alshamsi, Cappelli,
  Cojocaru, Debbah, Étienne Goffinet, Hesslow, Launay, Malartic, Mazzotta,
  Noune, Pannier, and Penedo}]{almazrouei2023falcon}
Ebtesam Almazrouei, Hamza Alobeidli, Abdulaziz Alshamsi, Alessandro Cappelli,
  Ruxandra Cojocaru, Mérouane Debbah, Étienne Goffinet, Daniel Hesslow,
  Julien Launay, Quentin Malartic, Daniele Mazzotta, Badreddine Noune, Baptiste
  Pannier, and Guilherme Penedo. 2023.
\newblock \href {http://arxiv.org/abs/2311.16867} {The falcon series of open
  language models}.

\bibitem[{Bai et~al.(2024)Bai, Kang, Stanovsky, Freitag, and
  Ritter}]{bai_schema-driven_2024}
Fan Bai, Junmo Kang, Gabriel Stanovsky, Dayne Freitag, and Alan Ritter. 2024.
\newblock \href {https://doi.org/10.48550/arXiv.2305.14336} {Schema-{Driven}
  {Information} {Extraction} from {Heterogeneous} {Tables}}.
\newblock ArXiv:2305.14336 [cs].

\bibitem[{Bingler et~al.(2022)Bingler, Kraus, Leippold, and
  Webersinke}]{bingler_how_2022}
Julia Bingler, Mathias Kraus, Markus Leippold, and Nicolas Webersinke. 2022.
\newblock \href {https://doi.org/10.2139/ssrn.4000708} {How {Cheap} {Talk} in
  {Climate} {Disclosures} relates to {Climate} {Initiatives}, {Corporate}
  {Emissions}, and {Reputation} {Risk}}.

\bibitem[{Borisov et~al.(2022)Borisov, Leemann, Sessler, Haug, Pawelczyk, and
  Kasneci}]{borisov_deep_2022}
Vadim Borisov, Tobias Leemann, Kathrin Sessler, Johannes Haug, Martin
  Pawelczyk, and Gjergji Kasneci. 2022.
\newblock \href {https://doi.org/10.1109/TNNLS.2022.3229161} {Deep {Neural}
  {Networks} and {Tabular} {Data}: {A} {Survey}}.
\newblock \emph{IEEE Transactions on Neural Networks and Learning Systems},
  pages 1--21.

\bibitem[{Brown et~al.(2020)Brown, Mann, Ryder, Subbiah, Kaplan, Dhariwal,
  Neelakantan, Shyam, Sastry, Askell, Agarwal, Herbert-Voss, Krueger, Henighan,
  Child, Ramesh, Ziegler, Wu, Winter, Hesse, Chen, Sigler, Litwin, Gray, Chess,
  Clark, Berner, McCandlish, Radford, Sutskever, and
  Amodei}]{brown_language_2020}
Tom Brown, Benjamin Mann, Nick Ryder, Melanie Subbiah, Jared~D Kaplan, Prafulla
  Dhariwal, Arvind Neelakantan, Pranav Shyam, Girish Sastry, Amanda Askell,
  Sandhini Agarwal, Ariel Herbert-Voss, Gretchen Krueger, Tom Henighan, Rewon
  Child, Aditya Ramesh, Daniel Ziegler, Jeffrey Wu, Clemens Winter, Chris
  Hesse, Mark Chen, Eric Sigler, Mateusz Litwin, Scott Gray, Benjamin Chess,
  Jack Clark, Christopher Berner, Sam McCandlish, Alec Radford, Ilya Sutskever,
  and Dario Amodei. 2020.
\newblock \href
  {https://proceedings.neurips.cc/paper/2020/hash/1457c0d6bfcb4967418bfb8ac142f64a-Abstract.html}
  {Language {Models} are {Few}-{Shot} {Learners}}.
\newblock In \emph{Advances in {Neural} {Information} {Processing} {Systems}},
  volume~33, pages 1877--1901. Curran Associates, Inc.

\bibitem[{Fang et~al.(2024)Fang, Xu, Tan, Zhang, Hu, Qi, Nickleach, Socolinsky,
  Sengamedu, and Faloutsos}]{fang_large_2024}
Xi~Fang, Weijie Xu, Fiona~Anting Tan, Jiani Zhang, Ziqing Hu, Yanjun Qi, Scott
  Nickleach, Diego Socolinsky, Srinivasan Sengamedu, and Christos Faloutsos.
  2024.
\newblock \href {https://doi.org/10.48550/arXiv.2402.17944} {Large {Language}
  {Models}({LLMs}) on {Tabular} {Data}: {Prediction}, {Generation}, and
  {Understanding} -- {A} {Survey}}.
\newblock ArXiv:2402.17944 [cs].

\bibitem[{Henisz et~al.(2019)Henisz, Koller, and Nuttall}]{henisz_five_2019}
Witold Henisz, Tim Koller, and Robin Nuttall. 2019.
\newblock \href
  {https://www.mckinsey.com/~/media/McKinsey/Business%20Functions/Strategy%20and%20Corporate%20Finance/Our%20Insights/Five%20ways%20that%20ESG%20creates%20value/Five-ways-that-ESG-creates-value.ashx}
  {Five ways that {ESG} creates value}.
\newblock \emph{McKinsey Quarterly}.

\bibitem[{Herzig et~al.(2020)Herzig, Nowak, M{\"u}ller, Piccinno, and
  Eisenschlos}]{herzig-etal-2020-tapas}
Jonathan Herzig, Pawel~Krzysztof Nowak, Thomas M{\"u}ller, Francesco Piccinno,
  and Julian Eisenschlos. 2020.
\newblock \href {https://doi.org/10.18653/v1/2020.acl-main.398} {{T}a{P}as:
  Weakly supervised table parsing via pre-training}.
\newblock In \emph{Proceedings of the 58th Annual Meeting of the Association
  for Computational Linguistics}, pages 4320--4333, Online. Association for
  Computational Linguistics.

\bibitem[{Jiang et~al.(2023)Jiang, Sablayrolles, Mensch, Bamford, Chaplot,
  Casas, Bressand, Lengyel, Lample, Saulnier, Lavaud, Lachaux, Stock, Scao,
  Lavril, Wang, Lacroix, and Sayed}]{jiang_mistral_2023}
Albert~Q. Jiang, Alexandre Sablayrolles, Arthur Mensch, Chris Bamford,
  Devendra~Singh Chaplot, Diego de~las Casas, Florian Bressand, Gianna Lengyel,
  Guillaume Lample, Lucile Saulnier, Lélio~Renard Lavaud, Marie-Anne Lachaux,
  Pierre Stock, Teven~Le Scao, Thibaut Lavril, Thomas Wang, Timothée Lacroix,
  and William~El Sayed. 2023.
\newblock \href {https://doi.org/10.48550/arXiv.2310.06825} {Mistral {7B}}.
\newblock ArXiv:2310.06825 [cs].

\bibitem[{Jiang et~al.(2024)Jiang, Sablayrolles, Roux, Mensch, Savary, Bamford,
  Chaplot, de~Las~Casas, Hanna, Bressand, Lengyel, Bour, Lample, Lavaud,
  Saulnier, Lachaux, Stock, Subramanian, Yang, Antoniak, Scao, Gervet, Lavril,
  Wang, Lacroix, and Sayed}]{Jiang2024MixtralOE}
Albert~Q. Jiang, Alexandre Sablayrolles, Antoine Roux, Arthur Mensch, Blanche
  Savary, Chris Bamford, Devendra~Singh Chaplot, Diego de~Las~Casas, Emma~Bou
  Hanna, Florian Bressand, Gianna Lengyel, Guillaume Bour, Guillaume Lample,
  L'elio~Renard Lavaud, Lucile Saulnier, Marie-Anne Lachaux, Pierre Stock,
  Sandeep Subramanian, Sophia Yang, Szymon Antoniak, Teven~Le Scao,
  Th{\'e}ophile Gervet, Thibaut Lavril, Thomas Wang, Timoth{\'e}e Lacroix, and
  William~El Sayed. 2024.
\newblock \href {https://api.semanticscholar.org/CorpusID:266844877} {Mixtral
  of experts}.
\newblock \emph{ArXiv}, abs/2401.04088.

\bibitem[{Kadra et~al.(2021)Kadra, Lindauer, Hutter, and
  Grabocka}]{kadra_well-tuned_2021}
Arlind Kadra, Marius Lindauer, Frank Hutter, and Josif Grabocka. 2021.
\newblock \href {https://openreview.net/forum?id=d3k38LTDCyO} {Well-tuned
  {Simple} {Nets} {Excel} on {Tabular} {Datasets}}.

\bibitem[{Kardas et~al.(2020)Kardas, Czapla, Stenetorp, Ruder, Riedel, Taylor,
  and Stojnic}]{kardas_axcell_2020}
Marcin Kardas, Piotr Czapla, Pontus Stenetorp, Sebastian Ruder, Sebastian
  Riedel, Ross Taylor, and Robert Stojnic. 2020.
\newblock \href {https://doi.org/10.18653/v1/2020.emnlp-main.692} {{AxCell}:
  {Automatic} {Extraction} of {Results} from {Machine} {Learning} {Papers}}.
\newblock In \emph{Proceedings of the 2020 {Conference} on {Empirical}
  {Methods} in {Natural} {Language} {Processing} ({EMNLP})}, pages 8580--8594,
  Online. Association for Computational Linguistics.

\bibitem[{{Levenshtein}(1966)}]{1966SPhD...10..707L}
V.~I. {Levenshtein}. 1966.
\newblock {Binary Codes Capable of Correcting Deletions, Insertions and
  Reversals}.
\newblock \emph{Soviet Physics Doklady}, 10:707.

\bibitem[{Liu et~al.(2021)Liu, Chen, Guo, Ziyadi, Lin, Chen, and
  Lou}]{liu_tapex_2021}
Qian Liu, Bei Chen, Jiaqi Guo, Morteza Ziyadi, Zeqi Lin, Weizhu Chen, and
  Jian-Guang Lou. 2021.
\newblock \href {https://openreview.net/forum?id=O50443AsCP} {{TAPEX}: {Table}
  {Pre}-training via {Learning} a {Neural} {SQL} {Executor}}.

\bibitem[{Liu et~al.(2020)Liu, Liu, Zhang, and Chen}]{Liu2020DNN2LRIF}
Zhaocheng Liu, Qiang Liu, Hao Zhang, and Yuntian Chen. 2020.
\newblock \href {https://api.semanticscholar.org/CorpusID:221266499} {Dnn2lr:
  Interpretation-inspired feature crossing for real-world tabular data}.
\newblock \emph{ArXiv}, abs/2008.09775.

\bibitem[{Lu et~al.(2022)Lu, Liu, Dai, Xiao, Lin, Han, Sun, and
  Wu}]{lu_unified_2022}
Yaojie Lu, Qing Liu, Dai Dai, Xinyan Xiao, Hongyu Lin, Xianpei Han, Le~Sun, and
  Hua Wu. 2022.
\newblock \href {https://doi.org/10.18653/v1/2022.acl-long.395} {Unified
  {Structure} {Generation} for {Universal} {Information} {Extraction}}.
\newblock In \emph{Proceedings of the 60th {Annual} {Meeting} of the
  {Association} for {Computational} {Linguistics} ({Volume} 1: {Long}
  {Papers})}, pages 5755--5772, Dublin, Ireland. Association for Computational
  Linguistics.

\bibitem[{Mishra et~al.(2024)Mishra, Berrospi, Dinkla, Antognini, Fusco,
  Bothur, Lysak, Livathinos, Nassar, Vagenas, Morin, Auer, Dolfi, and
  Staar}]{mishra_esg_2024}
Lokesh Mishra, Cesar Berrospi, Kasper Dinkla, Diego Antognini, Francesco Fusco,
  Benedikt Bothur, Maksym Lysak, Nikolaos Livathinos, Ahmed Nassar, Panagiotis
  Vagenas, Lucas Morin, Christoph Auer, Michele Dolfi, and Peter Staar. 2024.
\newblock \href {https://doi.org/10.1609/aaai.v38i21.30574} {{ESG}
  {Accountability} {Made} {Easy}: {DocQA} at {Your} {Service}}.
\newblock \emph{Proceedings of the AAAI Conference on Artificial Intelligence},
  38(21):23814--23816.
\newblock Number: 21.

\bibitem[{OpenAI et~al.(2023)OpenAI, Achiam, Adler, Agarwal, Ahmad, Akkaya,
  Aleman, Almeida, Altenschmidt, Altman, Anadkat, Avila, Babuschkin, Balaji,
  Balcom, Baltescu, Bao, Bavarian, Belgum, Bello, Berdine, Bernadett-Shapiro,
  Berner, Bogdonoff, Boiko, Boyd, Brakman, Brockman, Brooks, Brundage, Button,
  Cai, Campbell, Cann, Carey, Carlson, Carmichael, Chan, Chang, Chantzis, Chen,
  Chen, Chen, Chen, Chen, Chess, Cho, Chu, Chung, Cummings, Currier, Dai,
  Decareaux, Degry, Deutsch, Deville, Dhar, Dohan, Dowling, Dunning, Ecoffet,
  Eleti, Eloundou, Farhi, Fedus, Felix, Fishman, Forte, Fulford, Gao, Georges,
  Gibson, Goel, Gogineni, Goh, Gontijo-Lopes, Gordon, Grafstein, Gray, Greene,
  Gross, Gu, Guo, Hallacy, Han, Harris, He, Heaton, Heidecke, Hesse, Hickey,
  Hickey, Hoeschele, Houghton, Hsu, Hu, Hu, Huizinga, Jain, Jain, Jang, Jiang,
  Jiang, Jin, Jin, Jomoto, Jonn, Jun, Kaftan, Kaiser, Kamali, Kanitscheider,
  Keskar, Khan, Kilpatrick, Kim, Kim, Kim, Kirchner, Kiros, Knight, Kokotajlo,
  Kondraciuk, Kondrich, Konstantinidis, Kosic, Krueger, Kuo, Lampe, Lan, Lee,
  Leike, Leung, Levy, Li, Lim, Lin, Lin, Litwin, Lopez, Lowe, Lue, Makanju,
  Malfacini, Manning, Markov, Markovski, Martin, Mayer, Mayne, McGrew,
  McKinney, McLeavey, McMillan, McNeil, Medina, Mehta, Menick, Metz,
  Mishchenko, Mishkin, Monaco, Morikawa, Mossing, Mu, Murati, Murk, Mély,
  Nair, Nakano, Nayak, Neelakantan, Ngo, Noh, Ouyang, O'Keefe, Pachocki, Paino,
  Palermo, Pantuliano, Parascandolo, Parish, Parparita, Passos, Pavlov, Peng,
  Perelman, Peres, Petrov, Pinto, Michael, Pokorny, Pokrass, Pong, Powell,
  Power, Power, Proehl, Puri, Radford, Rae, Ramesh, Raymond, Real, Rimbach,
  Ross, Rotsted, Roussez, Ryder, Saltarelli, Sanders, Santurkar, Sastry,
  Schmidt, Schnurr, Schulman, Selsam, Sheppard, Sherbakov, Shieh, Shoker,
  Shyam, Sidor, Sigler, Simens, Sitkin, Slama, Sohl, Sokolowsky, Song,
  Staudacher, Such, Summers, Sutskever, Tang, Tezak, Thompson, Tillet,
  Tootoonchian, Tseng, Tuggle, Turley, Tworek, Uribe, Vallone, Vijayvergiya,
  Voss, Wainwright, Wang, Wang, Wang, Ward, Wei, Weinmann, Welihinda, Welinder,
  Weng, Weng, Wiethoff, Willner, Winter, Wolrich, Wong, Workman, Wu, Wu, Wu,
  Xiao, Xu, Yoo, Yu, Yuan, Zaremba, Zellers, Zhang, Zhang, Zhao, Zheng, Zhuang,
  Zhuk, and Zoph}]{openai_gpt-4_2023}
OpenAI, Josh Achiam, Steven Adler, Sandhini Agarwal, Lama Ahmad, Ilge Akkaya,
  Florencia~Leoni Aleman, Diogo Almeida, Janko Altenschmidt, Sam Altman,
  Shyamal Anadkat, Red Avila, Igor Babuschkin, Suchir Balaji, Valerie Balcom,
  Paul Baltescu, Haiming Bao, Mo~Bavarian, Jeff Belgum, Irwan Bello, Jake
  Berdine, Gabriel Bernadett-Shapiro, Christopher Berner, Lenny Bogdonoff, Oleg
  Boiko, Madelaine Boyd, Anna-Luisa Brakman, Greg Brockman, Tim Brooks, Miles
  Brundage, Kevin Button, Trevor Cai, Rosie Campbell, Andrew Cann, Brittany
  Carey, Chelsea Carlson, Rory Carmichael, Brooke Chan, Che Chang, Fotis
  Chantzis, Derek Chen, Sully Chen, Ruby Chen, Jason Chen, Mark Chen, Ben
  Chess, Chester Cho, Casey Chu, Hyung~Won Chung, Dave Cummings, Jeremiah
  Currier, Yunxing Dai, Cory Decareaux, Thomas Degry, Noah Deutsch, Damien
  Deville, Arka Dhar, David Dohan, Steve Dowling, Sheila Dunning, Adrien
  Ecoffet, Atty Eleti, Tyna Eloundou, David Farhi, Liam Fedus, Niko Felix,
  Simón~Posada Fishman, Juston Forte, Isabella Fulford, Leo Gao, Elie Georges,
  Christian Gibson, Vik Goel, Tarun Gogineni, Gabriel Goh, Rapha Gontijo-Lopes,
  Jonathan Gordon, Morgan Grafstein, Scott Gray, Ryan Greene, Joshua Gross,
  Shixiang~Shane Gu, Yufei Guo, Chris Hallacy, Jesse Han, Jeff Harris, Yuchen
  He, Mike Heaton, Johannes Heidecke, Chris Hesse, Alan Hickey, Wade Hickey,
  Peter Hoeschele, Brandon Houghton, Kenny Hsu, Shengli Hu, Xin Hu, Joost
  Huizinga, Shantanu Jain, Shawn Jain, Joanne Jang, Angela Jiang, Roger Jiang,
  Haozhun Jin, Denny Jin, Shino Jomoto, Billie Jonn, Heewoo Jun, Tomer Kaftan,
  Łukasz Kaiser, Ali Kamali, Ingmar Kanitscheider, Nitish~Shirish Keskar,
  Tabarak Khan, Logan Kilpatrick, Jong~Wook Kim, Christina Kim, Yongjik Kim,
  Hendrik Kirchner, Jamie Kiros, Matt Knight, Daniel Kokotajlo, Łukasz
  Kondraciuk, Andrew Kondrich, Aris Konstantinidis, Kyle Kosic, Gretchen
  Krueger, Vishal Kuo, Michael Lampe, Ikai Lan, Teddy Lee, Jan Leike, Jade
  Leung, Daniel Levy, Chak~Ming Li, Rachel Lim, Molly Lin, Stephanie Lin,
  Mateusz Litwin, Theresa Lopez, Ryan Lowe, Patricia Lue, Anna Makanju, Kim
  Malfacini, Sam Manning, Todor Markov, Yaniv Markovski, Bianca Martin, Katie
  Mayer, Andrew Mayne, Bob McGrew, Scott~Mayer McKinney, Christine McLeavey,
  Paul McMillan, Jake McNeil, David Medina, Aalok Mehta, Jacob Menick, Luke
  Metz, Andrey Mishchenko, Pamela Mishkin, Vinnie Monaco, Evan Morikawa, Daniel
  Mossing, Tong Mu, Mira Murati, Oleg Murk, David Mély, Ashvin Nair, Reiichiro
  Nakano, Rajeev Nayak, Arvind Neelakantan, Richard Ngo, Hyeonwoo Noh, Long
  Ouyang, Cullen O'Keefe, Jakub Pachocki, Alex Paino, Joe Palermo, Ashley
  Pantuliano, Giambattista Parascandolo, Joel Parish, Emy Parparita, Alex
  Passos, Mikhail Pavlov, Andrew Peng, Adam Perelman, Filipe de Avila~Belbute
  Peres, Michael Petrov, Henrique Ponde de~Oliveira Pinto, Michael, Pokorny,
  Michelle Pokrass, Vitchyr Pong, Tolly Powell, Alethea Power, Boris Power,
  Elizabeth Proehl, Raul Puri, Alec Radford, Jack Rae, Aditya Ramesh, Cameron
  Raymond, Francis Real, Kendra Rimbach, Carl Ross, Bob Rotsted, Henri Roussez,
  Nick Ryder, Mario Saltarelli, Ted Sanders, Shibani Santurkar, Girish Sastry,
  Heather Schmidt, David Schnurr, John Schulman, Daniel Selsam, Kyla Sheppard,
  Toki Sherbakov, Jessica Shieh, Sarah Shoker, Pranav Shyam, Szymon Sidor, Eric
  Sigler, Maddie Simens, Jordan Sitkin, Katarina Slama, Ian Sohl, Benjamin
  Sokolowsky, Yang Song, Natalie Staudacher, Felipe~Petroski Such, Natalie
  Summers, Ilya Sutskever, Jie Tang, Nikolas Tezak, Madeleine Thompson, Phil
  Tillet, Amin Tootoonchian, Elizabeth Tseng, Preston Tuggle, Nick Turley,
  Jerry Tworek, Juan Felipe~Cerón Uribe, Andrea Vallone, Arun Vijayvergiya,
  Chelsea Voss, Carroll Wainwright, Justin~Jay Wang, Alvin Wang, Ben Wang,
  Jonathan Ward, Jason Wei, C.~J. Weinmann, Akila Welihinda, Peter Welinder,
  Jiayi Weng, Lilian Weng, Matt Wiethoff, Dave Willner, Clemens Winter, Samuel
  Wolrich, Hannah Wong, Lauren Workman, Sherwin Wu, Jeff Wu, Michael Wu, Kai
  Xiao, Tao Xu, Sarah Yoo, Kevin Yu, Qiming Yuan, Wojciech Zaremba, Rowan
  Zellers, Chong Zhang, Marvin Zhang, Shengjia Zhao, Tianhao Zheng, Juntang
  Zhuang, William Zhuk, and Barret Zoph. 2023.
\newblock \href {https://doi.org/10.48550/arXiv.2303.08774} {{GPT}-4
  {Technical} {Report}}.
\newblock ArXiv:2303.08774 [cs].

\bibitem[{Paolini et~al.(2020)Paolini, Athiwaratkun, Krone, Ma, Achille,
  Anubhai, Santos, Xiang, and Soatto}]{paolini_structured_2020}
Giovanni Paolini, Ben Athiwaratkun, Jason Krone, Jie Ma, Alessandro Achille,
  Rishita Anubhai, Cicero Nogueira~dos Santos, Bing Xiang, and Stefano Soatto.
  2020.
\newblock \href {https://openreview.net/forum?id=US-TP-xnXI} {Structured
  {Prediction} as {Translation} between {Augmented} {Natural} {Languages}}.

\bibitem[{Pawlik and Augsten(2016)}]{pawlik_tree_2016}
Mateusz Pawlik and Nikolaus Augsten. 2016.
\newblock \href {https://doi.org/10.1016/j.is.2015.08.004} {Tree edit distance:
  {Robust} and memory-efficient}.
\newblock \emph{Information Systems}, 56:157--173.

\bibitem[{Raffel et~al.(2020)Raffel, Shazeer, Roberts, Lee, Narang, Matena,
  Zhou, Li, and Liu}]{raffel_exploring_2020}
Colin Raffel, Noam Shazeer, Adam Roberts, Katherine Lee, Sharan Narang, Michael
  Matena, Yanqi Zhou, Wei Li, and Peter~J. Liu. 2020.
\newblock \href {http://jmlr.org/papers/v21/20-074.html} {Exploring the
  {Limits} of {Transfer} {Learning} with a {Unified} {Text}-to-{Text}
  {Transformer}}.
\newblock \emph{Journal of Machine Learning Research}, 21(140):1--67.

\bibitem[{Schimanski et~al.(2024)Schimanski, Reding, Reding, Bingler, Kraus,
  and Leippold}]{schimanski_bridging_2024}
Tobias Schimanski, Andrin Reding, Nico Reding, Julia Bingler, Mathias Kraus,
  and Markus Leippold. 2024.
\newblock \href {https://doi.org/https://doi.org/10.1016/j.frl.2024.104979}
  {Bridging the gap in {ESG} measurement: {Using} {NLP} to quantify
  environmental, social, and governance communication}.
\newblock \emph{Finance Research Letters}, 61:104979.

\bibitem[{Schwarz et~al.(2017)Schwarz, Pawlik, and Augsten}]{schwarz_new_2017}
Stefan Schwarz, Mateusz Pawlik, and Nikolaus Augsten. 2017.
\newblock \href {https://doi.org/10.1007/978-3-319-68474-1_11} {A {New}
  {Perspective} on the {Tree} {Edit} {Distance}}.
\newblock In \emph{Similarity {Search} and {Applications}}, Lecture {Notes} in
  {Computer} {Science}, pages 156--170, Cham. Springer International
  Publishing.

\bibitem[{Sun et~al.(2019)Sun, Yang, Zhang, Lin, Dong, Young, and
  Dong}]{sun_supertml_2019}
Baohua Sun, Lin Yang, Wenhan Zhang, Michael Lin, Patrick Dong, Charles Young,
  and Jason Dong. 2019.
\newblock \href {https://doi.org/10.1109/CVPRW.2019.00360} {{SuperTML}:
  {Two}-{Dimensional} {Word} {Embedding} for the {Precognition} on {Structured}
  {Tabular} {Data}}.
\newblock In \emph{2019 {IEEE}/{CVF} {Conference} on {Computer} {Vision} and
  {Pattern} {Recognition} {Workshops} ({CVPRW})}, pages 2973--2981.
\newblock ISSN: 2160-7516.

\bibitem[{Touvron et~al.(2023)Touvron, Martin, and Stone}]{touvron_llama_2023}
Hugo Touvron, Louis Martin, and Kevin Stone. 2023.
\newblock \href
  {https://ai.meta.com/research/publications/llama-2-open-foundation-and-fine-tuned-chat-models/}
  {Llama 2: {Open} {Foundation} and {Fine}-{Tuned} {Chat} {Models}}.

\bibitem[{Wang et~al.(2021)Wang, Liu, Chen, Hong, Tang, and
  Song}]{wang_zero-shot_2021}
Chenguang Wang, Xiao Liu, Zui Chen, Haoyun Hong, Jie Tang, and Dawn Song. 2021.
\newblock \href {https://doi.org/10.18653/v1/2021.emnlp-main.94} {Zero-{Shot}
  {Information} {Extraction} as a {Unified} {Text}-to-{Triple} {Translation}}.
\newblock In \emph{Proceedings of the 2021 {Conference} on {Empirical}
  {Methods} in {Natural} {Language} {Processing}}, pages 1225--1238, Online and
  Punta Cana, Dominican Republic. Association for Computational Linguistics.

\bibitem[{Wang et~al.(2022{\natexlab{a}})Wang, Liu, Chen, Hong, Tang, and
  Song}]{wang_deepstruct_2022}
Chenguang Wang, Xiao Liu, Zui Chen, Haoyun Hong, Jie Tang, and Dawn Song.
  2022{\natexlab{a}}.
\newblock \href {https://doi.org/10.18653/v1/2022.findings-acl.67}
  {{DeepStruct}: {Pretraining} of {Language} {Models} for {Structure}
  {Prediction}}.
\newblock In \emph{Findings of the {Association} for {Computational}
  {Linguistics}: {ACL} 2022}, pages 803--823, Dublin, Ireland. Association for
  Computational Linguistics.

\bibitem[{Wang et~al.(2022{\natexlab{b}})Wang, Liu, and Song}]{wang_ielm_2022}
Chenguang Wang, Xiao Liu, and Dawn Song. 2022{\natexlab{b}}.
\newblock \href {https://doi.org/10.18653/v1/2022.emnlp-main.576} {{IELM}: {An}
  {Open} {Information} {Extraction} {Benchmark} for {Pre}-{Trained} {Language}
  {Models}}.
\newblock In \emph{Proceedings of the 2022 {Conference} on {Empirical}
  {Methods} in {Natural} {Language} {Processing}}, pages 8417--8437, Abu Dhabi,
  United Arab Emirates. Association for Computational Linguistics.

\bibitem[{Wang et~al.(2023)Wang, Zhou, Zu, Xia, Chen, Zhang, Zheng, Ye, Zhang,
  Gui, Kang, Yang, Li, and Du}]{wang_instructuie_2023}
Xiao Wang, Weikang Zhou, Can Zu, Han Xia, Tianze Chen, Yuansen Zhang, Rui
  Zheng, Junjie Ye, Qi~Zhang, Tao Gui, Jihua Kang, Jingsheng Yang, Siyuan Li,
  and Chunsai Du. 2023.
\newblock \href {https://doi.org/10.48550/arXiv.2304.08085} {{InstructUIE}:
  {Multi}-task {Instruction} {Tuning} for {Unified} {Information}
  {Extraction}}.
\newblock ArXiv:2304.08085 [cs].

\bibitem[{Wenhu~Chen and Wang(2020)}]{2019TabFact}
Jianshu Chen Yunkai Zhang Hong Wang Shiyang Li Xiyou~Zhou Wenhu~Chen,
  Hongmin~Wang and William~Yang Wang. 2020.
\newblock Tabfact : A large-scale dataset for table-based fact verification.
\newblock In \emph{International Conference on Learning Representations
  (ICLR)}, Addis Ababa, Ethiopia.

\bibitem[{Yin et~al.(2020)Yin, Neubig, Yih, and Riedel}]{yin-etal-2020-tabert}
Pengcheng Yin, Graham Neubig, Wen-tau Yih, and Sebastian Riedel. 2020.
\newblock \href {https://doi.org/10.18653/v1/2020.acl-main.745} {{T}a{BERT}:
  Pretraining for joint understanding of textual and tabular data}.
\newblock In \emph{Proceedings of the 58th Annual Meeting of the Association
  for Computational Linguistics}, pages 8413--8426, Online. Association for
  Computational Linguistics.

\bibitem[{Yu et~al.(2020)Yu, Wu, Lin, Wang, Tan, Yang, Radev, Socher, and
  Xiong}]{yu_grappa_2020}
Tao Yu, Chien-Sheng Wu, Xi~Victoria Lin, Bailin Wang, Yi~Chern Tan, Xinyi Yang,
  Dragomir Radev, Richard Socher, and Caiming Xiong. 2020.
\newblock \href {https://openreview.net/forum?id=kyaIeYj4zZ} {{GraPPa}:
  {Grammar}-{Augmented} {Pre}-{Training} for {Table} {Semantic} {Parsing}}.

\bibitem[{Zhu et~al.(2021)Zhu, Brettin, Xia, Partin, Shukla, Yoo, Evrard,
  Doroshow, and Stevens}]{zhu_converting_2021}
Yitan Zhu, Thomas Brettin, Fangfang Xia, Alexander Partin, Maulik Shukla,
  Hyunseung Yoo, Yvonne~A. Evrard, James~H. Doroshow, and Rick~L. Stevens.
  2021.
\newblock \href {https://doi.org/10.1038/s41598-021-90923-y} {Converting
  tabular data into images for deep learning with convolutional neural
  networks}.
\newblock \emph{Scientific Reports}, 11(1):11325.
\newblock Number: 1 Publisher: Nature Publishing Group.

\end{thebibliography}
